\documentclass[a4paper]{article}
\usepackage[text={7in,10in},centering]{geometry}
\usepackage{graphicx,color}
\usepackage[table, dvipsnames]{xcolor}
\usepackage{amsmath,amsthm,amssymb}
\usepackage{multirow,array}
\usepackage{tabularx}
\usepackage{caption,subcaption}
\usepackage{url}
\usepackage{hyperref}
\usepackage{cleveref}

\crefname{figure}{Figure}{Figure}
\Crefname{figure}{Figure}{Figure}
\crefname{table}{Table}{Table}
\Crefname{table}{Table}{Table}

\newcommand{\ones}{\mathbf 1}
\newcommand{\Prob}{\mathop{\bf prob}}
\newcommand{\reals}{{\mbox{\bf R}}}

\newcommand{\sign}{\mbox{\rm sign}}

\newcommand{\ci}{\mathrm{CI}} 
\newcommand{\iclr}{I_{\mathrm{clr}}}
\newcommand{\inwp}{I_{\mathrm{nwp}}}
\newcommand{\wm}{\mbox{W/m}^2} 

\graphicspath{{figures/}}

\title{Large width penalization for neural network-based prediction interval estimation\footnote{This work has been submitted to the IEEE for possible publication. Copyright may be transferred without notice, after which this version may no longer be accessible.} }
\author{Worachit Amnuaypongsa and Jitkomut Songsiri\footnote{Corresponding author} \\[1ex]
 Department of Electrical Engineering, Faculty of Engineering \\ [1ex]
 Chulalongkorn University, Bangkok, Thailand 10330 \\[1ex]
 e-mail: worachitam@gmail.com, and jitkomut.s@chula.ac.th
}   

\begin{document}
\maketitle

\begin{abstract}
Forecasting accuracy in highly uncertain environments is challenging due to the stochastic nature of systems. Deterministic forecasting provides only point estimates and cannot capture potential outcomes. Therefore, probabilistic forecasting has gained significant attention due to its ability to quantify uncertainty, where one of the approaches is to express it as a prediction interval (PI), that explicitly shows upper and lower bounds of predictions associated with a confidence level. High-quality PI is characterized by a high PI coverage probability (PICP) and a narrow PI width. In many real-world applications, the PI width is generally used in risk management to prepare resources that improve reliability and effectively manage uncertainty. A wider PI width results in higher costs for backup resources as decision-making processes often focus on the worst-case scenarios arising with large PI widths under extreme conditions. This study aims to reduce the large PI width from the PI estimation method by proposing a new PI loss function that penalizes the average of the large PI widths more heavily. The proposed formulation is compatible with gradient-based algorithms, the standard approach to training neural networks (NNs), and integrating state-of-the-art NNs and existing deep learning techniques. Experiments with the synthetic dataset reveal that our formulation significantly reduces the large PI width while effectively maintaining the PICP to achieve the desired probability. The practical implementation of our proposed loss function is demonstrated in solar irradiance forecasting, highlighting its effectiveness in minimizing the large PI width in data with high uncertainty and showcasing its compatibility with more complex neural network models. Therefore, reducing large PI widths from our method can lead to significant cost savings by over-allocation of reserve resources.
\end{abstract}


\section{Introduction}\label{sec:intro}
The rise in global temperatures due to climate change is a critical issue that demands global attention, prompting many nations to endorse a net zero carbon emissions policy \cite{Davis2018}. In accordance with the policy, many countries worldwide are promoting the production of renewable energy, including solar and wind energy, as demonstrated in \cite{Solangi2011, Zhang2013, Byrnes2013, Kitzing2012}. However, higher penetration of renewable energy sources into the power system can introduce significant uncertainty in power generation, which is highly dependent on natural factors and potentially impacts system reliability. To tackle reliability challenges, forecasting plays a crucial role today in predicting the future value of clean energy. This forecast assists in shaping future generation preparation plans by utilizing the predicted values as inputs for unit commitment and economic dispatch problem \cite{Li2020review}. The traditional forecast, deterministic forecast, provides only a single value that does not reflect the uncertainty of forecasting. Recently, a probabilistic forecast has gained a lot of attention for its ability to provide the uncertainty information of the forecasted value \cite{Gneiting2014}. The probabilistic forecast can represent the uncertainty in three forms: quantile, distribution function, and prediction interval (PI), which can help the user in decision-making in various applications, especially in power applications such as solar \cite{reviewvandermeer2018, Li2020review}, wind \cite{Zhang2014}, electrical load \cite{Hong2016, Zhao2020}, electricity price \cite{Khosravi2013price, Nowotarski2018} forecasting.
The PI is widely used for uncertainty quantification because it effectively illustrates possible outcomes by displaying the upper and lower bounds associated with the confidence level \cite{Quan2020}. For example, in \cite{Apostolopoulou2018}, the uncertainty of solar generation represented by the PI was incorporated into a robust optimization for the economic dispatch problem of a hydroelectric system. In \cite{Cordova2018}, a prediction interval and the central value of renewable generation were included in the unit commitment problem to determine the resources needed for risk hedging.

\paragraph{Prediction interval construction approach.} 
There are indirect and direct approaches to PI construction. \emph{The indirect method} requires two steps: first, train a point forecast model, and second, conduct a statistical analysis of the forecast error. For instance, in \cite{Lorenz2009}, the authors generated a point forecast for solar irradiance based on a physical model, subsequently assuming that the error followed a normal distribution to calculate the PI. In \cite{Li2022}, XGBoost was employed for a deterministic solar irradiance forecast, and the kernel density estimation was then applied to the forecast error to construct the PI. For the neural network-based approach, methods to provide prediction intervals (PI) include the delta method, Bayesian method, and mean-variance estimation (MVE) method, as summarized in \cite{Khosravi2011}. The delta and Bayesian methods can incur high computational costs due to the need for calculating the Jacobian and Hessian matrices, respectively \cite{Quan2020}. On the other hand, the MVE method tends to perform poorly when the target variable does not follow the Gaussian assumption. Thus, the distribution of forecast error relates to the model, meaning the performance of the PI relies on both the model and the statistical analysis method.
On the other hand, \emph{the direct approach} offers PI from a model in one step, as the model is specifically trained to account for the effect of the input on the uncertainty characteristics of the target variable \cite{Quan2020}. Quantile-based techniques, such as quantile regression (QR) \cite{Koenker2005} and quantile regression forest (QRF) \cite{Meinshausen2006}, can be applied to construct PIs without requiring distribution assumption. Among direct methods, the lower upper bound estimation (LUBE) technique is favorable as it directly outputs the upper and lower bounds from the NN, originally proposed in \cite{Khosravi2011}. This study focuses on the direct method for constructing PI, where the model is directly trained with a loss function defined by PI quality. The quality of PI is evaluated based on reliability and sharpness, commonly measured through prediction interval coverage probability (PICP) and PI width, respectively. A high-quality PI is characterized by a high PICP and a narrow PI width. However, these two goals conflict, presenting a trade-off characteristic where enhancing PICP results in a wider PI width. To achieve a high PICP and a narrow PI width, a common strategy is to introduce a loss function that effectively represents the quality of the PIs for both aspects.

\paragraph{Prediction interval-based loss function.} 
Many literatures formulate the PI construction problem as an optimization problem in various ways. The loss function can merge the PICP and PI width components into a scalar-valued loss function. The LUBE approach introduces a coverage width-based criterion (CWC) as a loss function for training the NN, utilizing the multiplicative form for the PICP and PI width functions \cite{Khosravi2011}. Several alternative versions of CWC loss have been proposed within the LUBE framework, utilizing various models such as support vector machine (SVM) \cite{Shrivastava2015}, extreme learning machine (ELM) \cite{Ni2017}, and NN \cite{Quan2014, Ye2016}, with NN being the most widely used. In \cite{Quan2014constrained}, the authors defined the problem as minimizing the PI width, treating PICP as a constraint to achieve the desired probability. However, imposing the PICP as a hard constraint resulted in a larger PI width than necessary. Consequently, \cite{Zhang2015} introduced a deviation information-based criterion (DIC) as a new objective in an additive form that incorporated both PI width and the \emph{pun} function, which penalizes the deviation of PI from the target variable. Schemes that treated the problem as a multi-objective optimization challenge, aiming to minimize the PI width and maximize the PICP as bi-objectives, were proposed in \cite{Li2018mo, Galvan2017mo}. However, all of these problem formulations utilized heuristic optimization techniques such as simulated annealing and particle swarm optimization to find the optimal model parameters, which did not guarantee a local minimum of the loss function. Additionally, the loss function is incompatible with gradient-based algorithms due to its non-differentiability, while gradient-based methods are the standard for training state-of-the-art NN \cite{Chen2024}.

\paragraph{PI-loss function compatible with gradient-based algorithm.} 
To develop a loss function suitable for the gradient-based method, \cite{Pearce2018} introduced the quality-based (QD) loss function, which includes the PICP and PI width terms, relying on statistical principles from the likelihood framework. The authors implemented a smooth approximation of the count function in PICP to ensure differentiability. Additionally, there are several enhanced versions of QD loss. For instance, \cite{Salem2020, Lai2022} refined the QD loss by incorporating a mean squared error (MSE) term to deliver point forecasts alongside the PI from ensemble NN models. In \cite{Saeed2024}, the authors presented an improved QD version by introducing a calibration function that imposes greater penalties on uncovered PIs to enhance multi-horizon predictive capabilities. This improved QD approach was utilized with a gated multi-scale convolutional sequence model. A multi-objective framework was available for the gradient-based approach, as detailed in \cite{Chen2024}, which used the multi-gradient descent algorithm (MGDA) \cite{MOGD} along with the QD loss to identify the optimal descent direction for two objectives. Therefore, the loss compatible with the gradient-based algorithm leads to further implementation of more complex state-of-the-art NN models such as Long Short-Term Memory (LSTM) and the Transformers. As these methods perform well in constructing PI, the resulting PI usually has the optimal sense of average PI width because the average PI width is typically part of a loss function. 

\paragraph{Cost of large PI widths.} 
For example, in power system applications, information about PI width is utilized to prepare reserve resources, addressing potential uncertainties. A wider PI indicates greater uncertainty in the forecast, which may require a larger reserve margin to ensure reliability, resulting in higher costs due to an increased need for backup resources. Especially in power system contexts, the PI of renewable forecasting is generally eployed to quantify the uncertainty in renewable generation, directly assisting in the decision-making process. These intervals help power operators handle uncertainty, allowing them to plan reserve power generation to meet future electrical demand and increase the stability of the power system. In \cite{Zhao2021}, the authors demonstrated that the deviation of point forecasts from the PIs in wind power applications was used to determine reserve power, impacting operational costs. According to \cite{Zhao2022}, the PI of wind power was utilized to determine the amount of offered wind power within the PI, where the decision-making regarding the offered wind power was based on the worst-case cost. However, a large PI width, which may occur in some instances due to high uncertainty, can drastically affect the worst-case cost, impacting decision-making. In practice, unit commitment and economic dispatch often rely on robust optimization problems to address the uncertainty \cite{Quan2020}. The robust approach requires defining an uncertainty set to account for variability in renewable generation, emphasizing preparation for the worst-case scenario under extreme conditions, especially in cases with extremely large PI widths. As a result, the solution from robust optimization is generally considered conservative. The PI widths are generally a mix of small and large values, but a portion of a large PI width that occurs in some instances can lead to more reserve power preparation at all times. Therefore, merely focusing on the average PI width may not be sufficient. It is also essential to monitor the group of large PI widths because planning under a significant number of large PI widths can raise operational costs. When the large PI width is reduced, the uncertainty set in the robust optimization scheme becomes smaller, which can be advantageous in reducing the conservatism of the unit commitment solution, as the worst-case scenario is less severe. A smaller uncertainty set can lead to reduced operating costs and enhanced efficiency while maintaining reliability since the robust optimization approach can focus on more realistic scenarios without the overallocation of reserve resources.

\paragraph{Contribution.}
This study directly aims to provide a reduction in large PI width. In previous work, PI loss functions typically evaluate PI width as the average width. However, relying solely on the average PI width may not effectively capture cases of large PI widths. In \cite{Amnuaypongsa2024}, the authors introduced a PI construction framework using a convex optimization formulation to control the large and maximum PI widths, aiming to reduce extreme values using a linear additive model. In this study, we propose a loss function with a new PI width function that penalizes the large PI width more severely. The large PI width can be reduced while the PICP lies within the desired coverage probability. Additionally, our proposed loss function is compatible with gradient-based optimization algorithms, allowing advanced neural network models to be integrated into the PI estimation framework. This enhances the linear additive model from \cite{Amnuaypongsa2024} to more effectively capture the nonlinear characteristics found in real-world data, leveraging the capabilities of advanced deep learning techniques. The contribution of our work can be applied to any application where the large PI width affects the decision-making, not only in the power system. For example, reducing the large PI width in landslide displacement prediction can avoid overestimating the uncertainty in highly volatile data due to various stochastic natural features, which improves risk planning and potentially lowers the costs of developing a risk action plan \cite{Lian2016}. The PI of crop yield prediction in agriculture is used to provide strategies for uncertainty management to maximize profit \cite{Morales2023}. Decision-making includes resource allocation plans, such as nitrogen fertilizer rates, which involve significant costs. Reducing the large PI width not only lowers the costs when decisions are based on worst-case scenarios but also minimizes environmental impact by reducing resource wastage, particularly in highly uncertain areas. In construction projects, reducing the large PI width in material costs can significantly enhance the efficiency of financial resource allocation, ensuring better control over budgets and reducing financial risks where the best- and worst-case scenarios are usually considered in project management \cite{Mir2021}.

This article is structured as follows. \Cref{sec:background} outlines the background of PI estimation, the relevant PI evaluation metrics, and the LUBE method. \Cref{sec:methodology} details our methodology for providing PI, including our design of the loss function and the numerical methods used. \Cref{sec:experimentsetting} describes how the experiments are designed to compare performance against benchmark methods. The advantages of our formulation on synthetic data are shown in \Cref{sec:exponsyntheticdata}. Lastly, \Cref{sec:exponrealdataset} shows an experiment on real-world applications, focusing on solar irradiance forecasting.

\section{Background of PI estimation} \label{sec:background}
Given a dataset $\mathcal{D} = \{x_{i}, y_{i}\}_{i=1}^{N}$ with $N$ samples where $x \in \reals^{p}$ represented a $p$-dimensional predictor vector and $y \in \reals^n$ denoted an $n$-dimensional target variable. The target variable is generated by a data-generating process (DGP) described as $y = f(x) + e$, where $f(x)$ is the ground truth function and $e$ is an additive data noise. A regression task aims to predict the value of $y$ given the predictors $x$. A regression model estimates the relationship between $y$ and $x$, parametrized by the model parameters $\theta$, expressed as $\hat{f}(x; \theta)$. Therefore, the overall uncertainty of $y$ is composed of model uncertainty and data noise. A prediction interval (PI) is a statistical tool that quantifies the overall uncertainty of $y$ by providing an upper and lower bound of $y$ at a confidence level of $(1-\delta) \times 100\%$. In the direct PI construction approach, a model directly estimates the relationship between the PI and $x$ using the model parameter $\theta$, such that $\hat{f}(x;\theta) = \left( \hat{l}, \hat{u} \right)$, where $\hat{l} \triangleq \hat{l}(x; \theta)$ and $\hat{u} \triangleq \hat{u}(x; \theta)$ represent the estimated lower and upper bound of the PI respectively. A PI estimation method seeks to construct an interval such that the probability of covering $y$ equals $1-\delta$ shown as
	\begin{equation}
		\Prob\left(\hat{l}(x; \theta) \leq y \leq \hat{u}(x; \theta)\right) = 1 - \delta.
	\end{equation}
As the confidence level $1-\delta$ increases, the PI inherently widens to capture more data points, ensuring higher coverage probability.
\subsection{PI evaluation metrics}

The performance of PI is assessed based on reliability and sharpness that can be quantified from a high PICP and narrow PI width, respectively. However, these two objectives exhibit a trade-off relationship, necessitating a complete evaluation of the PI performance that considers both PICP and PI width. This section outlines the fundamental evaluation metrics for PI. We denote that the sample width of the PI is given by $w_{i} = \hat{u}_{i} - \hat{l}_{i}$ used in the study.

\begin{enumerate}
\item \textbf{Prediction interval coverage probability (PICP)} is the measure of reliability calculated by counting the proportion of the observers lying within the PIs.  The PICP can be defined as
\begin{equation}\label{PICP}
\text{PICP} = \frac{1}{N}\sum_{i=1}^{N} \ones (\hat{l}_{i}\leq y_{i} \leq \hat{u}_{i}),
\end{equation}
where $\ones(A)$ is a count function that returns one if event $A$ occurs and zero otherwise. The PICP is expected to reach the confidence level set for PI estimation.
	
\item \textbf{Prediction interval width} measures the sharpness of the prediction interval that can be assessed in terms of averaged PI width across all samples called prediction interval average width $\text{PIAW} = \frac{1}{N}\sum_{i=1}^{N} w_{i}$. Additionally, there is a normalized version of the PIAW called prediction interval normalized average width (PINAW), which eliminates the scale of the target variable shown below:
		\begin{equation}\label{PINAW}
			\text{PINAW} = \frac{1}{NR}\sum_{i=1}^{N} w_{i},
		\end{equation}
where $R = y_{\text{max}} - y_{\text{min}}$ indicates the target variable's range, computed as the difference between the maximum and minimum values of the target variable, ensuring the PINAW scale is between 0 and 1.
	
	\item \textbf{Winkler score} \cite{winkler1972} is a metric that evaluates the reliability and sharpness of the PIs corresponding to the $(1-\delta) \times 100 \%$ coverage probability. The Winkler score is equivalent to a pinball loss when evaluating PIs with equal-tailed probability. A low Winkler score indicates that the lower and upper bounds of the PI match well with the quantiles of $\frac{\delta}{2}$ and $1 - \frac{\delta}{2}$. The Winkler score is commonly found in the literature related to PI-based methods, while the pinball loss is predominantly found in quantile-based approaches. This study utilizes the Winkler score in a normalized version, expressed as:
    		\begin{equation}\label{Winklerscore}
			\text{Winkler} = \frac{1}{NR} \sum^{N}_{i=1} \vert \hat{u}_{i}-\hat{l}_{i} \vert + \frac{2}{\delta} \left [ (\hat{l}_{i}-y_{i}) \ones(y_{i} < \hat{l}_{i}) + (y_{i} - \hat{u}_{i}) \ones(y_{i} > \hat{u}_{i})\right], 
		\end{equation}
where the normalization factor $R$ is used to eliminate data scaling effects in the Winkler score.
\end{enumerate}

\subsection{LUBE Method}\label{sec:LUBEmethod}

\begin{figure}[ht]
		\centering
			\includegraphics[width=0.45\linewidth]{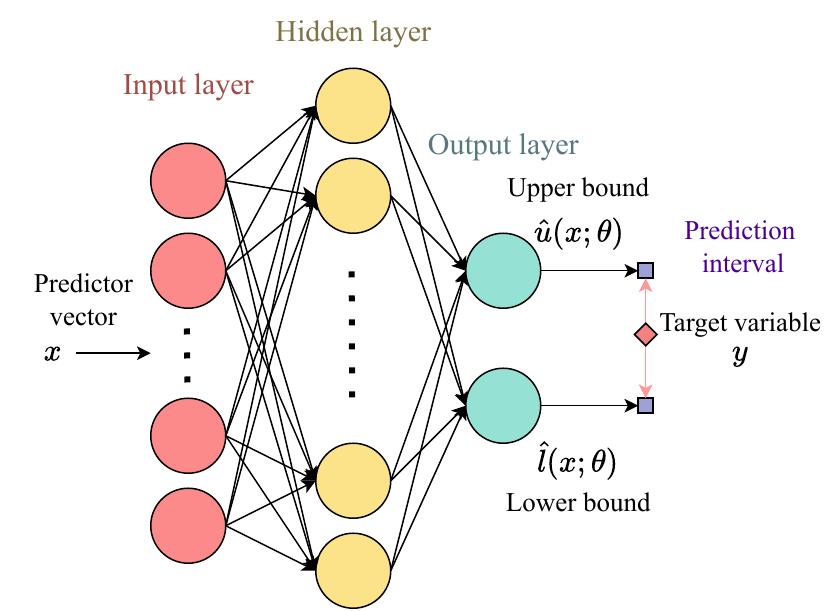}
			\caption{The NN model with two outputs utilized in LUBE framework.}
		\label{fig:lubemethod}
\end{figure}

A lower upper bound estimation (LUBE) is a technique that provides a PI directly from a model without a distribution assumption for the target variable. The model directly maps the input features $x$ to the upper and lower bounds to capture an uncertainty of $y$. The literature presents various models that establish PIs, including SVM \cite{Shrivastava2015}, ELM \cite{Zhang2015}, \cite{Zhao2020}, \cite{Zhao2022}, and neural networks (NN), where the latter has been the most widely used option. Typically, the NN-based model is implemented with two output neurons representing the upper and lower bounds, as illustrated in \cref{fig:lubemethod}. The PI with the $1-\delta$ confidence level is obtained by minimizing the PI-based loss function. The PI-based loss function has two objectives: PICP and PI width, which can be combined either multiplicatively or additively into a single loss function. The PI-based loss function was originally introduced as the coverage-width criterion (CWC) loss in \cite{Khosravi2011}. The CWC function integrates both objectives using the following multiplicative form:
\begin{equation}\label{CWCori}
	\text{CWC}_{\text{ori}}(\theta) = \text{PINAW}(\theta)\left( 1 + e^{\gamma \max(0, (1-\delta) - \text{PICP}(\theta))} \right),
\end{equation}
where $\gamma$ balances the trade-off between the two objectives. The alternative versions of CWC loss are also represented in the multiplicative form as noted in \cite{Quan2014, Ye2016}. However, studies \cite{Shrivastava2015, Pearce2018} mention that the global minimum for \eqref{CWCori} occurs when the PINAW has a PI width of zero, which is commonly an undesirable solution. Following this, additive versions of CWC are alternatively suggested to address the multiplicative drawback identified in the studies by \cite{Shrivastava2015, Shi2018}. 

The PICP term requires computing a count function, which results in a non-differentiable loss. As a result, a population-based algorithm, which is a heuristic approach, is usually applied to identify the optimal model parameter. Recently, the log-likelihood-based loss in an additive form called a quality-driven (QD) loss was proposed in \cite{Pearce2018}. The PI width term in the QD loss is calculated by averaging only the samples with $y_{i}$ captured by the PI as: $\text{PIAW}_\text{capt.} = \frac{1}{N_{\text{capt.}}} \sum_{i=1}^{N} w_{i}\ones(\hat{l}_{i} \leq y_{i} \leq \hat{u}_{i})$ where $N_{\text{capt.}} = \sum_{i=1}^{N}\ones(\hat{l}_{i} \leq y_{i} \leq \hat{u}_{i})$. This guarantees that the model does not gain advantages from the reduced PI width in the sample that is not covered by the PI. For the PICP term, with i.i.d. assumption for all samples, a value of the count function $\ones(\hat{l}_{i} \leq y_{i} \leq \hat{u}_{i})$ can be treated as a Bernoulli random variable with probability $1-\delta$. Consequently, the number of covered samples $N_{\text{capt.}}$ is modeled as $\text{Binomial}(N, (1-\delta))$. The central limit theorem allows for approximating the probability mass function of the binomial distribution as a normal distribution when $N$ is sufficiently large: $\text{Binomial}(N, (1-\delta)) \approx \mathcal{N}\left( N(1-\delta), N\delta(1-\delta) \right)$. Hence, the negative log-likelihood of $N_{\text{capt.}}$ denoted as $\mathcal{L}_{\text{PICP}}$, can be approximated as:

\begin{align}
	-\log \mathcal{L}_{\text{PICP}} & \propto  \frac{N}{\delta(1-\delta)}((1-\delta)-\text{PICP})^{2}.
\end{align}
Finally, the $\max(0, \cdot)$ is equipped to penalize only when $\text{PICP} < 1-\delta$, resulting in the QD loss expressed as: 
\begin{equation}\label{lossqd_ori} 
		\text{Loss}_{\text{QD}}(\theta) = \text{PIAW}_{\text{capt.}}(\theta) + \gamma \frac{N}{\delta (1-\delta)}\text{max}(0, (1-\delta)-\text{PICP}(\theta))^{2},
\end{equation}
where $\gamma$ controls a trade-off level between two objectives. Furthermore, in \cite{Pearce2018}, a smooth approximation of the count function is introduced in the PICP term to ensure that the QD loss is differentiable and that gradient-based algorithms can be applied for optimization, which will be detailed in \cref{sec:smoothapprox}.

\section{Methodology} \label{sec:methodology}
\begin{figure}[ht]
	\centering
		\includegraphics[width=0.5\linewidth]{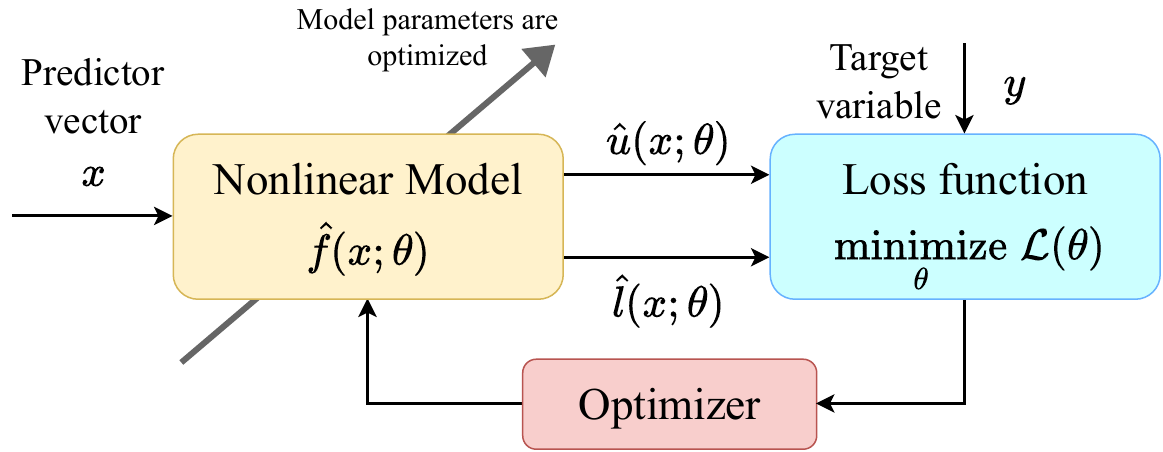}
		\caption{The methodology of training mechanism of the PI construction.}
	\centering
	\label{fig:pimethodology}
\end{figure}

\cref{fig:pimethodology} outlines the training procedures for PI construction. There are three key components: a model, a loss function, and an optimizer. First, the model refers to any nonlinear model that receives $x$ as inputs and produces $\hat{u}(x;\theta)$ and $\hat{l}(x;\theta)$, both defined by the trainable parameter $\theta$. Second, the loss function assesses the quality of PI, where a lower loss reflects better PI performance. The loss function evaluates the value using PI information from $\hat{u}, \hat{l}$, and $y$. Third, an optimizer is a numerical method for solving the proposed loss function. The closed loop in the \cref{fig:pimethodology} represents the iterative process of updating the model parameters, $\theta$, to minimize the loss function until the stopping criterion is met. After completing the training process, the model with the optimized parameters can generate the PIs where the PICP achieves the confidence level. In this study, we are concerned with introducing a novel PI-based loss function, the sum of $k$-largest component loss, which incorporates a new aspect of PI width penalization. To illustrate the idea in this paper, we utilize a feedforward neural network (ANN) model as shown in \cref{fig:lubemethod}.

\subsection{Sum-$k$ formulation}

We design a loss function to provide PIs that reduce the \emph{large} PI width while ensuring the PICP achieves the desired coverage probability ($1 - \delta$). To reduce the large PI widths, our loss function is specifically designed to impose a high penalty on large PI widths relative to narrow PI widths. The large PI widths typically arise when data are corrupted by heteroskedastic noise dependent on $x$, resulting in higher uncertainty in certain regions of $x$. We aim to reduce the excessive PI width in these regions.

The proposed loss function comprises two components: the PICP function and the PI width function, in the additive form similar to the QD loss. Our \textbf{sum-$k$ loss function} can be expressed as
\begin{equation} \label{sumkform}
	\mathcal{L}_{\text{sum}-k}(\theta) = \max(0, (1-\delta) - \text{PICP}(\theta)) + \gamma \mathcal{W}(\theta),
\end{equation}
where $\gamma>0$ controls the trade-off between the PICP and the PI width. Increasing $\gamma$ penalizes more heavily on the PI width in the loss function, resulting in a narrower PI width while losing the PICP. 

Considering the PICP term in the loss function, the PICP term is designed to obtain the lowest deviation between PICP and $1 - \delta$, aiming to penalize only when the PICP falls below $1 - \delta$. So, we utilize the $\max(0, \cdot)$ function in the PICP term, following \cite{Pearce2018}. In addition, we modify the quadratic function in the PICP term in QD loss to a linear function, as shown in \eqref{sumkform}. Our proposed formulation imposes a stronger penalty on PICP deviation than QD loss when the PICP is slightly near the desired probability. This is because a linear function penalizes more heavily than a quadratic function within a small range where $x>x^{2}$ for $x \in (0, 1)$. The $\text{PICP}$ can be calculated according to the definition in \eqref{PICP}.

Next, considering the PI width term, $\mathcal{W}(\theta)$, we propose a new PI width function employing the sum of the $K$-largest function to impose a strong penalty on large PI widths, referred to in our loss function as the \emph{sum-$k$} loss ($\mathcal{L}_{\text{sum}-k}$), expressed in \eqref{sumkform}. The PI width function has two terms: the average of the $K$-largest PI widths and the average of the remaining PI widths shown as
\begin{equation} \label{sumkwidth}
\mathcal{W}(\theta) := \mathcal{W}(\theta | K, \lambda) = \frac{1}{R_{\text{quantile}}} \left [ \frac{1}{K} \sum_{i=1}^{K} w_{[i]}(\theta) + \lambda \cdot \frac{1}{N-K} \sum_{K+1}^{N} w_{[i]}(\theta) \right ], 
\end{equation}
where $w_{[i]}$ is the $i^{\mathrm{th}}$ largest PI width element, with $w_{[1]} \geq w_{[2]} \geq \cdots \geq w_{[N]}$. In \eqref{sumkwidth}, $K$ is the number of samples treated as large PI widths, while the others are considered narrow PI widths. A hyperparameter, $\lambda>0$, represents a relative weight of the averaged narrow PI width to the averaged large PI width. When $\lambda < 1$, the average of the $K$-largest PI widths is penalized more heavily than the others. The hyperparameter $K$ is set to $\lfloor kN \rfloor$ where $k \in (0, 1)$ is a portion of the data that will be treated as large PI widths. A higher value of $k$ refers to a greater number of PI widths samples, categorized as large widths during the training process. In addition, a new normalization factor, quantile range ($R_{\text{quantile}}$), is also utilized to eliminate the effect of outliers and the scale of PI width in \eqref{sumkwidth}, which is expressed as
\begin{equation}
	R_{\text{quantile}} = q_{y}(0.95) - q_{y}(0.05),
\end{equation}
where $q_{y}(0.95)$ and $q_{y}(0.05)$ are the corresponding quantiles of $y$ at 0.95 and 0.05 probability of $y$. The proposed PI width function is a special case of the ordered weighted $\ell_{1}$ norm, which has been shown to be a convex function and classified as a norm in the studies by \cite{Bogdan2013, Zeng2014, Zeng2015, Figueiredo2016}.

\begin{figure}[ht]
	\centering
		\includegraphics[width=1\linewidth]{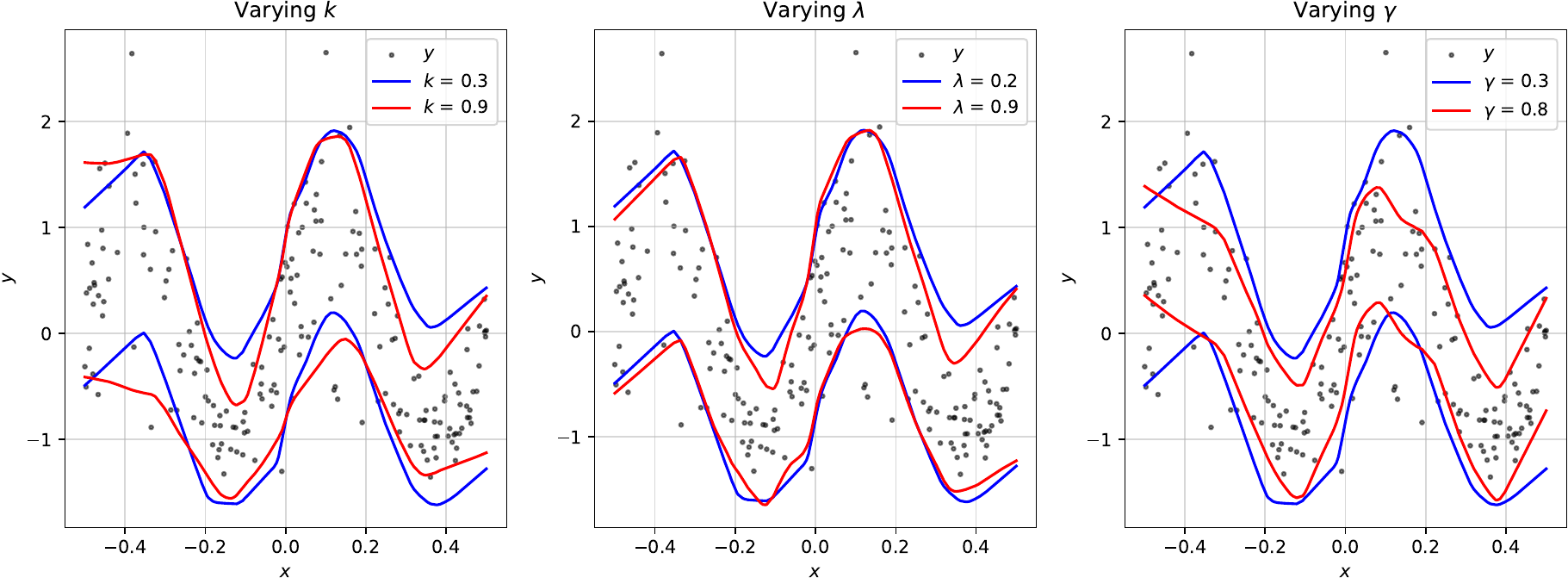}
		\caption{The effect of the hyperparameter in the sum-$k$ formulation: (left) Reducing $k$ decreases large PI widths. (central) Lowering $\lambda$ reduces large PI widths while increasing narrow widths. (right) Increasing $\gamma$ reduces in overall PI width but loosing the PICP.}
		
	\centering
	\label{fig:sumk_hypereffect}
\end{figure}

Therefore, the sum-$k$ formulation has three hyperparameters that need to be tuned: $k, \lambda, \gamma$. For $k$, as it is close to one, the amount of PI width categorized as the large PI width increases. Setting lower $k$ highlights the difference penalization between the large and narrow PI width, causing a heavy reduction in the large PI width. The value of $k$ should be determined by examining the proportion of high-variance samples in the dataset. As a start, users can set $k = 0.3$ in practice. The hyperparameter $\lambda$ controls the relative penalization level of the narrow PI width compared to the large PI width. Reducing $\lambda$ leads to a decrease in the large PI width; the narrow PI widths are wider since the loss function penalizes less on the narrow PI width. We recommend setting $\lambda$ based on the user's preferences. For $\gamma$ in \eqref{sumkform}, different PICP levels are achievable by adjusting $\gamma$, where a higher $\gamma$ reduces the PI width but decreases PICP. The value of $\gamma$ consistently remains within the tenth decimal scale, independent of sample sizes and data scale, because the loss function is already normalized with $N$ and $R_{\text{quantile}}$, making it easier to tune $\gamma$ than QD loss in \cite{Pearce2018}. To choose an appropriate $\gamma$, we recommend tuning $\gamma$ based on cross-validated PICP. Start by splitting the data into training and validation sets for the training process. Then, vary $\gamma$ and choose the value that achieves the desired PICP at the confidence level on the validation set. This approach ensures that the model’s performance is evaluated on an unseen dataset, leading to better generalization. In summary, $k$ and $\lambda$ are required to be set according to user preferences, while $\gamma$ should be adjusted using the validation set. The impact of adjusting each hyperparameter is illustrated in \cref{fig:sumk_hypereffect}.

\subsection{Smooth approximation of a count function} \label{sec:smoothapprox}
\begin{figure}[ht]
	\centering
		\includegraphics[width=0.7\linewidth]{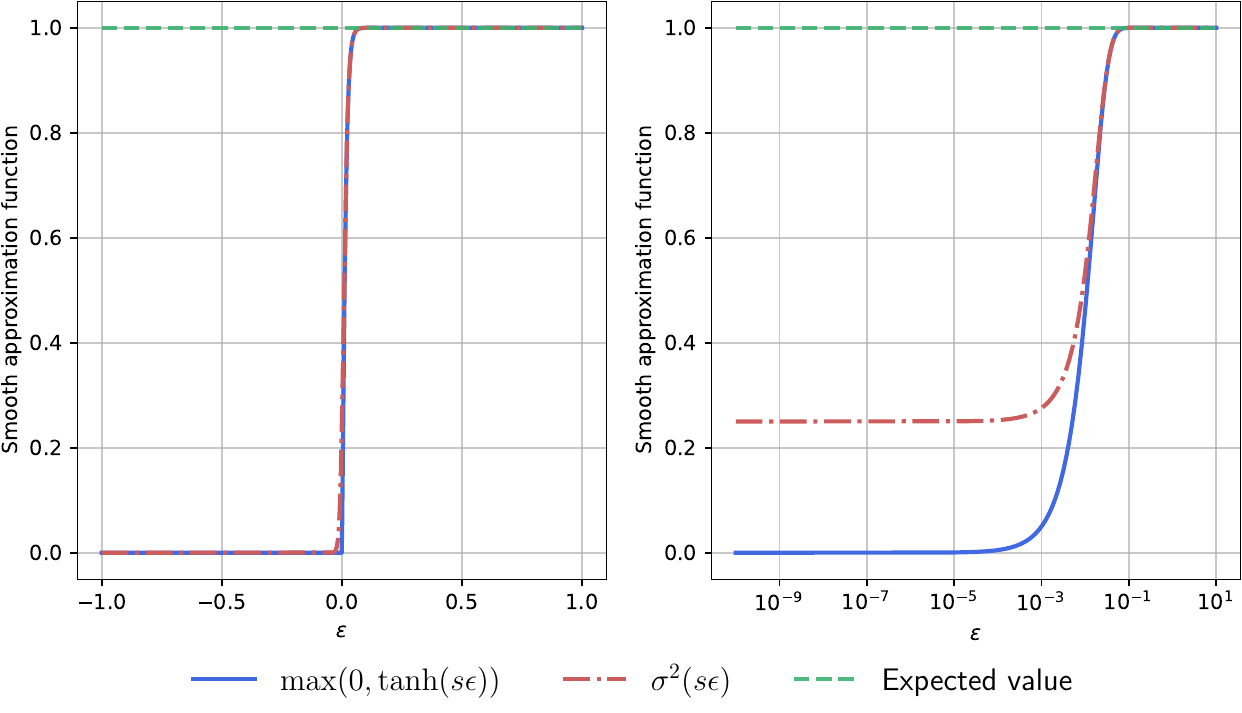}
		\caption{The margin effect of the smooth approximation: (left) Comparison of smooth approximation function values. (right) Impact of the margin on the smooth approximation displayed on a log scale.}
	\centering
	\label{fig:smoothapprox}
\end{figure}

The calculation of PICP as defined in \eqref{PICP} in the sum-$k$ formulation involves a count function, which is non-differentiable, making it unsuitable for gradient-based algorithms. To address this issue, in \cite{Pearce2018, Lai2022, Chen2023, Saeed2024}, a smooth approximation of the count function is presented as a product of sigmoid functions.
\begin{equation} \label{sigmoidapprox}
	 \ones_{\text{sigmoid}}(\hat{l}_{i}(\theta) \leq y_{i} \leq \hat{u}_{i}(\theta)) =
	 \sigma (s(y_{i} - \hat{l}_{i}(\theta)) ) \cdot
	 \sigma(s(\hat{u}_{i}(\theta) - y_{i}) ),
\end{equation}
where $\sigma(x) = \frac{1}{1+e^{-x}}$ and $s$ is the softening factor. In this study, we introduce an alternative smooth approximation using the sum of the hyperbolic tangent functions, called the $\tanh$-smooth approximation, which is simpler than \eqref{sigmoidapprox}. 
To show the simplicity of the $\tanh$-smooth approximation, Firstly, we define a $\tanh$-count function $\xi$ calculated as
\begin{equation} \label{tanhapprox}
	\xi_{i} =  \frac{1}{2} \left( \tanh(s(y_{i} - \hat{l}_{i}(\theta))) + \tanh(s(\hat{u}_{i}(\theta) - y_{i} )) \right), 
\end{equation}
where $\xi_{i} \rightarrow 1$ when $y_{i} \in \left( \hat{l}_{i}, \hat{u}_{i} \right)$. From $\sigma(x) = \frac{1+\tanh\left(\frac{x}{2}\right)}{2}$, we can show that
\begin{align} \label{tanhsigmoidprove}
	\sigma(s(y_{i} - \hat{l}_{i}(\theta)) ) \cdot
	 \sigma(s(\hat{u}_{i}(\theta) - y_{i}) ) & = \frac{1}{4} \left(1+\tanh \left( \frac{s}{2}(y_{i}-\hat{l}_{i}(\theta)) \right) \right) \left(1+\tanh \left( \frac{s}{2}(\hat{u}_{i}(\theta) - y_{i}) \right) \right) \nonumber \\
	 & = \frac{1}{4}\left[ 1 + \tanh \left( \frac{s}{2}(y_{i}-\hat{l}_{i}(\theta)) \right)\tanh \left( \frac{s}{2}(\hat{u}_{i}(\theta) - y_{i}) \right) + 2 \xi_{i} \right] \nonumber. 
\end{align}
It can be seen that $\xi$ is part of the sigmoid approximation, showing a simpler mathematical expression of the $\tanh$-smooth approximation. Moreover, to obtain a similar curve from the two functions, it is recommended to set the softening factor $s$ of the $\tanh$ function to be half that of the sigmoid function. However, $\xi$ can be negative since the $\tanh$ function ranges from $(-1, 1)$. The negative case occurs when the upper bound lies below and the lower bound lies above the observation, respectively, which is undesirable. To prevent misbehavior of the counting function, the operation $\max(0, \cdot)$ is equipped in $\xi$ to set the counting function to zero in this scenario. Therefore, the proposed $\tanh$-smooth approximation can be expressed as
\begin{equation} \label{tanhapproxadjusted}
	 \ones_{\text{tanh}}(\hat{l}_{i}(\theta) \leq y_{i} \leq \hat{u}_{i}(\theta)) =
	 \frac{1}{2} \max \left (0, \tanh(s(y_{i} - \hat{l}_{i}(\theta))) + \tanh(s(\hat{u}_{i}(\theta) - y_{i} ))\right ).
\end{equation}
The $\tanh$-smooth approximation utilizes the $\max(0, x)$ function, which is non-smooth due to its non-differentiability at $x=0$. Nevertheless, most NN software implementations provide one of the derivatives on either side of zero instead of generating an error \cite{Goodfellow-et-al-2016}. This allows $\max(0, x)$ function to stay compatible with gradient-based algorithms \cite{Pearce2018}. 
In the case of the $\tanh$-smooth approximation, the case that $\xi$ is exactly zero is rarely encountered numerically, allowing us to consider it as a smooth function. In addition, employing any choice of smooth approximations leads to a numerical problem when $\hat{u}_{i}, \hat{l}_{i}, y_{i}$ are exactly equal. As indicated in \eqref{sigmoidapprox} and \eqref{tanhapprox}, the smooth approximations do not yield the value of one in such cases, meaning that this sample is not qualified as a PI-covered sample. Therefore, to be considered a PI-covered sample, a small margin between $\hat{u}_{i}, \hat{l}_{i}, y_{i}$ is necessary. Suppose $\hat{u}_{i} = y_{i} + \epsilon$ and $\hat{l}_{i} = y_{i}+\epsilon$, the sigmoid and $\tanh$-smooth approximation now can be simplified to $\sigma^{2}(s\epsilon), \max(0, \tanh(s\epsilon))$ respectively. \cref{fig:smoothapprox} shows the differences in the choices of smooth approximation (left) and the effect of margin on the smooth approximation (right). The left panel in \cref{fig:smoothapprox} shows that the $\tanh$-smooth approximation closely resembles the sigmoid approximation when the softening factor is set to 50 and 100, respectively. The right panel in \cref{fig:smoothapprox} shows the smooth approximation function varying with margin $\epsilon$, indicating a minimum margin of approximately 0.1 for a PI-covered sample. The minimum margin decreases with a higher softening factor, but excessive softening can cause a divergence in the model training process due to backpropagation. In this study, the PICP within the loss function employs the $\tanh$-smooth approximation function shown as:
\begin{equation}
	\text{PICP}(\theta) = \frac{1}{N} \sum_{i=1}^{N} \ones_{\text{tanh}}(\hat{l}_{i}(\theta) \leq y_{i} \leq \hat{u}_{i}(\theta)),
\end{equation}
where $s$ is set to 50, demonstrating good convergence results for the gradient-based algorithm.

\subsection{Numerical method}
A numerical problem in this study is to minimize the proposed loss function to obtain optimal model parameters, which is an optimization problem. As the sum-$k$ loss function involves a sum of nonlinear smooth approximations in the PICP term, the overall loss function is highly nonlinear. The optimization problem can be cast as an unconstrained nonlinear optimization problem. In addition, the sum-$k$ loss is continuous and differentiable, allowing the use of a gradient-based algorithm to solve the problem. The benefit of using a gradient-based optimization algorithm is that it enables the application of various techniques, such as minibatch optimization, adaptive learning rates for faster convergence, momentum-based updates to help escape from flat regions, and improved stability overall.  For simple linear models (but the loss function is still nonlinear in parameters), stochastic gradient descent and AdaGrad \cite{duchi2011} can be implemented to minimize the sum-$k$ loss. In the case of large models, often found in NN, the number of model parameters is usually very high. Several widely used optimization tools are available for training NNs, including RMSProp \cite{Tieleman2012}, Adam \cite{Kingma2017}, and Nadam \cite{dozat2016}. These optimizers enhance the training of large NN by using adaptive learning rates and a momentum mechanism, resulting in faster and more stable convergence on nonlinear and high-dimensional loss surfaces.

\section{Experiment setting} \label{sec:experimentsetting}
This section describes the setting of all experiments, including benchmarked methods, evaluation metrics, and the experiment design. The codes and synthetic datasets utilized in this article are accessible at \url{https://github.com/energyCUEE/PIestim_sumk}.

\subsection{Evaluation metrics} \label{sec:evalmetrics}
This section describes all evaluation metrics employed to assess the performance of PI in the experiments. The reliability of PI is quantified by the PICP \eqref{PICP}. The sharpness of PI is represented by the average PI width (PINAW) \eqref{PINAW}. Additionally, the Winkler score \eqref{Winklerscore} is employed to illustrate the correspondence of the PI derived from each method with the equal-tail quantile. For PI width and Winkler score evaluation, a new normalization factor called the quantile range $R_{\text{quantile}} = q_{y}(0.95) - q_{y}(0.05)$ calculated from the target variable's quantiles at 0.05 and 0.95 probabilities, is implemented to eliminate the influence of outliers in the target variable. The $R_{\text{quantile}}$ is utilized in place of $R$ in \eqref{PINAW} and \eqref{Winklerscore} throughout the entire study. As this study focuses on reducing the large PI width, a metric called the Prediction Interval Normalized Average Large Width (PINALW) has been developed. The PINALW is the average of normalized \emph{large} PI width calculated from the sample that has a PI width more than the $p$-quantile of PI width.
	\begin{equation}\label{PINALW}
		\text{PINALW}(p) = \frac{1}{KR_{\text{quantile}}}\sum_{i=1}^{K} w_{[i]},
	\end{equation}
where $w_{[i]}$ represents the $i^{\mathrm{th}}$ largest element of the PI width such that $w_{[1]} \geq w_{[2]} \geq \cdots \geq w_{[N]}$, and $K = \lfloor (1 - p)N \rfloor$ is the number of samples classified as having large PI widths corresponding to the $p$-quantile. In this study, we choose the PINALW with $p = 0.5$ to evaluate the average width of the top half of the PI width, referring to the large PI width. 

\subsection{Benchmarked methods}
Various methods are selected to construct the PI and compare the performance of PIs from our formulation. Most of the chosen methods are in the form of the loss function, which can be minimized with a NN model. The trainable parameter of the NN model is denoted as $\theta$. The formulation with combined terms between PICP and PI width is modified in the same format as ours, where the penalty parameter $\gamma$ is multiplied to the PI width penalty function.

\begin{enumerate}
	\item \textbf{Mean-variance estimation (MVE)} estimates the mean and variance of the target variable, relying on the Gaussian assumption \cite{Nix1994}. Our study applies the MVE to the neural network with two output neurons: mean ($\hat{\mu}(x_{i};\theta)$) and variance ($\hat{\sigma}^{2}(x_{i};\theta)$).  The loss function of MVE is based on the log-likelihood of the Gaussian distribution as:
		\begin{equation}\label{lossmve}
			\text{Loss}_{\text{MVE}}(\theta) = \frac{1}{2} \sum_{i=1}^{N} \left (  \log(\hat{\sigma}^{2}(x_{i};\theta)) + \frac{(y_{i} - \hat{\mu}(x_{i};\theta))^{2}}{\hat{\sigma}^{2}(x_{i}; \theta)} \right).
		\end{equation}
After estimating $\hat{\mu}(x_{i};\theta)$ and $\hat{\sigma}^{2}(x_{i};\theta)$, the PI is constructed based on the given confidence level as $\hat{\mu}(x_{i};\theta) \pm z_{1 - \frac{\delta}{2}}\hat{\sigma}(x_{i}, \theta)$, where $z_{1-\frac{\delta}{2}}$ is the $z$-score corresponding with a $(1-\delta) \times 100\%$  confidence level.

%		\begin{equation}\label{pimve}
%			\hat{\mu}(x_{i};\theta) \pm z_{1 - \frac{\delta}{2}}\hat{\sigma}(x_{i}, \theta),
%		\end{equation}

	\item \textbf{Quantile regression (QR)} is the method that estimates the conditional quantile of the target variable ($y$) given the predictor ($x$) \cite{Koenker2005}. The quantile value is determined by minimizing the pinball loss for a specified quantile. To construct a PI with a confidence level of $(1-\delta)\times 100 \%$ from QR, we define the lower and upper bounds based on the quantiles $\frac{\delta}{2}$ and $1 - \frac{\delta}{2}$, respectively. The pinball loss functions associated with these quantiles are combined as the loss function for training the neural network (NN) with two output nodes. 
		\begin{equation}\label{pinballloss}
			\text{Loss}_\text{Pinball}(\theta) = \frac{1}{N} \left [ \sum_{i=1}^{N} \rho_{\frac{\delta}{2}}(y_{i}-\hat{l}(x_{i};\theta)) +  \rho_{1-\frac{\delta}{2}}(y_{i}-\hat{u}(x_{i};\theta)) \right ],
		\end{equation}
where $\rho_{\alpha}(r) = \max(\alpha r, (\alpha - 1)r)$ is a pinball function.
	\item \textbf{Quantile regression forest (QRF)} is a tree-based method using the random forest framework to provide the full conditional CDF \cite{Meinshausen2006}. The estimated CDF is determined by calculating the average of the indicator function of the target variable across all decision trees. To determine a PI, two quantiles are chosen to define the upper and lower bounds at the specified confidence level, similar to QR.
	\item \textbf{Coverage-width-based criterion (CWC)} is a PI-based loss function that directly combines the PICP and PI width term as a single loss function. The CWC was first proposed in \cite{Khosravi2011} as a loss function that combines PINAW and PICP for training NN of a lower upper bound estimation (LUBE) method. The LUBE approaches have many alternative loss functions. In \cite{Quan2014}, the PINAW in the CWC loss is replaced by PINRW shown as
		\begin{equation}\label{CWCquan}
				\text{CWC}_{\text{Quan}}(\theta) = \text{PINRW}( 1 +  \ones(\text{PICP} < 1 - \delta) \cdot e^{-\gamma(\text{PICP} - (1 - \delta) )}).
		\end{equation}
		The PINRW utilized the 2-norm concept for the PI width term, which penalizes more on the large width calculated as $\text{PINRW} = \frac{1}{R_{\text{quantile}}} \sqrt{\frac{1}{N} \sum_{i=1}^{N} (\hat{u}(x_{i};\theta)-\hat{l}(x_{i};\theta))^{2}}$. In this study, we adjust \eqref{CWCquan} to its equivalence, \eqref{CWCquaneq}, to ensure continuity, allowing it to be solved with a gradient-based algorithm as the benchmark method.
		\begin{equation}\label{CWCquaneq}
				\text{CWC}_{\text{Quan-eq}}(\theta) = \text{PINRW}(1 +  e^{\gamma \text{max}(0, (1 - \delta) - \text{PICP})}).
		\end{equation}
		However, the multiplicative term of PI width in \eqref{CWCquaneq} can lead to abnormal characteristics of PI width where the PINRW reaches zero, as derived from the global minimum of the loss function. Subsequently, the authors of \cite{Shrivastava2015} modified the CWC loss function to be an additive form as demonstrated in \eqref{CWCshri}.
		\begin{equation}\label{CWCshri}
		            \text{CWC}_{\text{Shri}}(\theta) = \text{PINAW} +   \ones(\text{PICP} < 1 - \delta) \cdot e^{-\gamma(\text{PICP}-(1-\delta))}.
		\end{equation}
	 The version presented in \eqref{CWCshrieq} is implemented in our experiment to guarantee continuity and compatibility with gradient-based algorithms.
			\begin{equation}\label{CWCshrieq}
				\text{CWC}_{\text{Shri-eq}}(\theta) = \text{PINAW} + e^{\gamma \text{max}(0, (1 - \delta) - \text{PICP})}.
		\end{equation}
		In \cite{Li2020}, the authors presented a CWC formulation \eqref{CWCli} to evaluate PI more efficiently when the PICP is below the desired coverage. They apply an affine transformation to the PI width term of the CWC function. Because the original CWC is notably sensitive to the PICP term, the $\text{CWC}_{\text{Li}}$ is proposed to address the issue that the original CWC fails to accurately evaluate the variation of the PINAW when the PICP is less than $1 - \delta$.	
		\begin{equation}\label{CWCli}
				\text{CWC}_{\text{Li}}(\theta) = 
				\begin{cases}
					\beta \text{PINAW}, & \text{PICP} \geq 1 - \delta \\
					(\alpha + \beta \text{PINAW})( 1 + e^{-\gamma(\text{PICP}-(1-\delta))}),  & \text{PICP} < 1 - \delta
				\end{cases}
		\end{equation}
		However, the $\text{CWC}_{\text{Li}}$ is not continuous in the PICP argument. Therefore, we also modify it to \eqref{CWClicont} as a continuous version to be applicable to a gradient-based algorithm.
		\begin{equation}\label{CWClicont}
				\text{CWC}_{\text{Li-eq}}(\theta) = \frac{\beta}{2} \text{PINAW} + \left( \alpha + \frac{\beta}{2} \text{PINAW} \right ) e^{\gamma \text{max}(0, (1 - \delta) - \text{PICP})}.
		\end{equation}
	\item \textbf{Deviation information-based criterion (DIC)} is proposed in \cite{Zhang2015} to account for the PIs deviation information in the loss function shown in \eqref{DIC}. The exponential term in the CWC is substituted with the function $\mathrm{pun}$, as defined in \eqref{pun}, within the DIC loss framework. This adjustment is made to evaluate the deviation of the target variable from the PI based on samples lying outside the PI.
	
		\begin{equation}\label{DIC}
			\text{DIC}(\theta) = \text{PINAW} + \ones(\text{PICP} < 1 - \delta) \cdot \mathrm{pun}
		\end{equation}
		where 
		\begin{equation}\label{pun}
			\mathrm{pun} = \gamma \left [  \sum_{i=1}^{N_{L}} (\hat{l}(x_{i};\theta) - y_{i}) +  \sum_{i=1}^{N_{U}} (y_{i} - \hat{u}(x_{i};\theta) )     \right ],
		\end{equation}
	where $N_{L}$ and $N_{U}$ represent the number of observations located below $\hat{l}(x_{i}, \theta)$ or above $\hat{u}(x_{i}, \theta)$, respectively. The penalty parameter $\gamma$ is set as $1/\delta$ according to \cite{Zhang2015}.
	\item \textbf{Quality driven loss function (QD)} is proposed based on the high-quality principle of obtaining narrow PI with achieving a desired PICP \cite{Pearce2018}. The QD loss consists of two components: PI width and coverage probability, combined in the additive form as presented in \eqref{lossqd_ori}.
	
The PI width term is proposed as the captured PI width measured from only the PI that captures the data point. The coverage term is derived based on the likelihood principle. The count function is approximated using a smooth version, and then the loss function is proposed as applicable to the gradient-based algorithm. In this study, we modify the original QD loss by adjusting the trade-off parameter $\gamma$ to focus on penalizing the PI width term. The influences of $N$ and $\delta$ in \eqref{lossqd_ori} are removed, and instead, the PINAW is employed to enable $\gamma$ to manage the trade-off regardless of the number of samples, the desired probability, and the data range. As a result, the updated version used in this study, presented in \eqref{lossqdmod}, is still equivalent to \eqref{lossqd_ori}.
		\begin{equation}\label{lossqdmod}
		\text{Loss}_{\text{QD-eq}}(\theta) = \text{max}(0, (1-\delta)-\text{PICP})^{2} + \gamma \text{PINAW}_{\text{capt.}}.
			\end{equation}
\end{enumerate}

\subsection{Experiment design} \label{subsec:expdesign}
This section explains how we designed the experiments in this study. The experiments are divided into three parts: the first two utilize four synthetic datasets, each containing 100 noise trials, while the last dataset is from a real-world solar energy application. The first experiment explores the trade-off characteristics between the PICP and the PI width, including PINAW and PINALW from each formulation. The second experiment compares the characteristics of the PI width and the performance indices from each method with a controlled equal PICP. The third experiment applied the formulations in solar forecasting applications, which exhibit a high level of uncertainty, to demonstrate the performance of each formulation in real-world contexts. The desired coverage probability ($1-\delta$) was set at 0.9 across all experiments. The $\tanh$ smooth approximation with $s=50$ was utilized in the methods involving the PICP term in the loss function. The design of all experiments is detailed below.

\subsubsection{Trade-off characteristics} \label{subsec:tradeoffexp}
This experiment aims to compare the trade-off characteristics among the methods with formulation hyperparameters that control the trade-off between PICP and the PI width indices, including QD, $\text{CWC}_{\text{Quan}}$, $\text{CWC}_{\text{Shri}}$, $\text{CWC}_{\text{Li}}$, and Sum-$k$ loss. The penalty hyperparameters $ \gamma $ were linearly spaced between $ \gamma_{\text{min}} $ and $ \gamma_{\text{max}} $, with ten distinct values to reach the PICP range from 0.85 to 0.9 for each dataset. For $\text{CWC}_{\text{Li}}$, the hyperparameters $\alpha$ and $\beta$ were set according to the original paper. For the sum-$k$ loss, we initially set $k = 0.3$ to impose a greater penalty on the widest $30\%$ of PI widths while using $\lambda = 0.1$ to highlight the penalty differences between narrow and large PI widths. Trade-off curves were generated by varying the formulation hyperparameter and calculating the PICP and PI width for each data trial. Finally, the PICP and PI width from all trials for each dataset were averaged to generalize the results, allowing for plotting the trade-off curve.

An ANN model was controlled with the same architecture across all methods shown as \cref{tab:modelhyperparameters} and set the random seed to control the initialization of model parameters. For the optimizer, the Adam optimizer was selected to minimize the loss due to its efficiency in avoiding stuck in local minima and its computational efficiency \cite{Kingma2017}. In this study, the learning rate was selected based on the characteristics of the loss function and was adjusted according to the observed loss to achieve fast convergence. The batch size was set based on the number of samples used. In the training algorithm, the maximum number of epochs was set as 2,000 to terminate the training process. We set the patience to 100, representing the maximum number of epochs without loss function improvement, which serves as an early stopping criterion. The trade-off curves are evaluated on the validation set.

\begin{table}[ht]
	\centering
	\caption{The model architecture of ANN utilized in the synthetic data experiment.}
	\begin{tabular}{| c | c |}
		\hline
		\textbf{Model specification} & \textbf{Setting} \\
		\hline
		\multicolumn{1}{|l|}{Hidden layers} &  3   \\
		\hline
		\multicolumn{1}{|l|}{Neurons per layer} &  no. of input features, 100, 100, 100, 2  \\
		\hline
		\multicolumn{1}{|l|}{Activation function} & ReLU \\
		\hline
		\multicolumn{1}{|l|}{Batch Normalization} & Added after hidden layers \\
		\hline
		\multicolumn{1}{|l|}{Total number of trainable parameters} & 21,102 + no. of input features $\times$ 100 \\
		\hline
	\end{tabular}
	\label{tab:modelhyperparameters}
\end{table}

\subsubsection{Width characteristics with control PICP in validation set} \label{subsubsec:expdesign_controlpicp} 
This experiment compared the performance metrics mentioned in \cref{sec:evalmetrics} and the PI width distribution of all methods with a controlled PICP of 0.9, with all evaluations conducted on the validation set. The methods used in the comparison are QR, QRF, MVE, DIC, $\text{CWC}_{\text{Quan}}$, $\text{CWC}_{\text{Shri}}$, $\text{CWC}_{\text{Li}}$, and QD. 
For methods that contain formulation hyperparameters: $\text{CWC}_{\text{Quan}}$, $\text{CWC}_{\text{Shri}}$, $\text{CWC}_{\text{Li}}$, QD, and our formulation, we selected the operating point from the trade-off curve experiment where PICP of validation set is closest to 0.9 for each dataset for comparison. For MVE, QR, QRF, and DIC, a confidence level of 0.9 is the only required hyperparameter. For QR and QRF, we set the lower and upper bound quantiles as 0.05 and 0.95, respectively. The performance metrics were presented as the mean and standard deviation calculated from various noise trials. At the same time, the distribution of PI width for each formulation was illustrated as a histogram plot aggregated from all 100 data trials.

\subsubsection{Real world data: solar irradiance forecasting}
Handling uncertainty in solar irradiance forecasting presents notable challenges due to varying weather conditions. Furthermore, solar irradiance often fluctuates significantly due to unpredictable cloud cover. This application is a suitable illustration of the data characterized by heteroskedastic noise and high volatility, particularly in Thailand, which is positioned near the equator. Therefore, the experiment was designed to demonstrate a reduction in the large PI widths observed during high volatility periods of the proposed formulation compared to other methods. For this experiment, we used the same ANN models with our formulation and benchmark methods, including QR, QD, and $\text{CWC}_\text{Shri}$. We also applied the LSTM model with our formulation to demonstrate its performance across different NN architectures. The results include a comparison of performance indices and characteristics of the solar irradiance PI, shown in a time series plot with varying sky conditions. We also described characteristics of forecasted PI across different horizons to illustrate when PI was released in real operation.

\section{Experiment on synthetic data} \label{sec:exponsyntheticdata}
The performance of all methods is demonstrated using four synthetic datasets containing heteroskedastic noise. In total, 100 noise trials with consistent ground truth are generated to conclude the generalizability of the results. The data is randomly split $80\%$ for training and $20\%$ for validation. The descriptions of the datasets are provided below.

\subsection*{Dataset}
The dataset contains four synthesis data with different mathematical functions. The DGP of all datasets is in the form of $y = f(x) + e$ where $f(x)$ is the ground truth function and $e$ is the corrupted noise.
\begin{enumerate}
	\item \textbf{Sum of Gaussian function:} $f(x) = \beta_{0} + \sum_{i=1}^{4} \beta_{i} \exp - \frac{(x - \mu_{i})^{2}}{2}$ where $x$ is generated as a uniform distribution in $[-4, 4]$ with 2,000 samples. $\beta_{0}, \cdots, \beta_{4}$ are one-time generated from $\mathcal{N}(1,1)$ while $\mu_{1}, \cdots, \mu_{4}$ are equal to -2.4, -0.8, 0.8, and 2.4 respectively. The noise is generated from $\mathcal{N}\left ( 0, (\sqrt{2\max(0, \sign(\vert x \vert - 1.5)}) + 0.2)^{2} \right )$. The resulting DGP contains both high and low volatile noise regions.
	
	\item \textbf{Polynomial function} \cite{pmlr-v37-hernandez-lobatoc15}, \cite{Lin2021}: $f(x) = x^{3}$ where $x$ is randomly generated as $\mathcal{U}(-4, 4)$ with 1,000 samples. The function is applied with heteroskedastic noise instead of uniform variance noise in the original papers. The noise $e$ is generated from $\mathcal{N}(0, (2\vert x \vert + \exp(x))^{2})$.

	\item \textbf{Sinusoid function} \cite{JMLR:v22:20-1100}: For this dataset, $y$ is generated from $\mathcal{N}(\sin(4\pi x), (0.5 + 0.3\sin(4 \pi x))^{2})$ with 1,000 samples where $x \in [-0.5, 0.5]$.
	
	\item \textbf{Multivariate function} \cite{Papadopoulos2001}: $f(x) = 10\sin(\pi x_{1}x_{2}) + 20(x_{3} - 0.5)^{2} + 10x_{4} + 5x_{5}$ where each $x_{i}$ is randomly generated as $\mathcal{U}(0, 1)$ with 1,000 samples. Each noise samples $e$ is generated with heteroskedasticity from $\mathcal{N}(0, 9\Vert x \Vert_{2}^{2})$.
\end{enumerate}

\subsection*{Result and discussion}
	\begin{figure}[ht]
		\centering
			\includegraphics[width=0.9\linewidth]{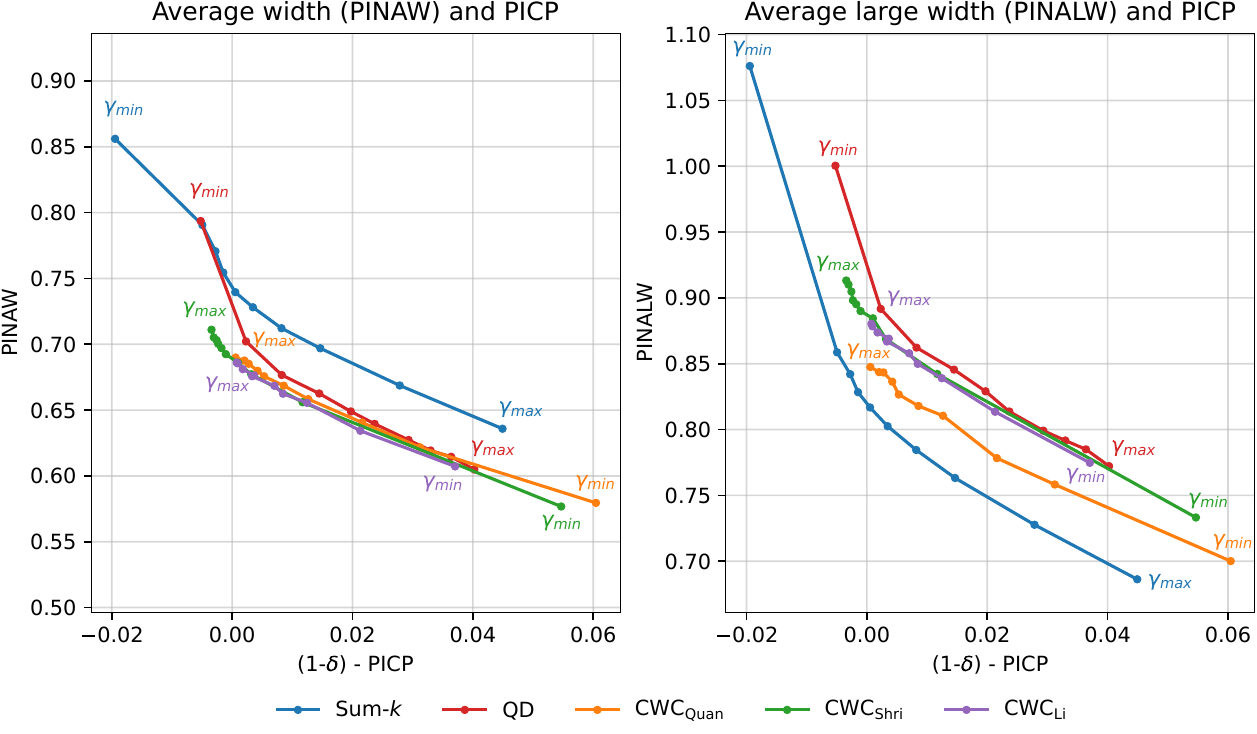}
			\caption{Comparison of the trade-off curve by varying the trade-off parameter for each formulation in the sum of Gaussian dataset. (left): Between $(1-\delta)-\text{PICP}$ and PINAW. (right): Between $(1-\delta)-\text{PICP}$ and PINALW.}
		\centering
		\label{fig:tradeoff_dgpgaussian}
	\end{figure}
\paragraph{Trade-off characteristics.} The trade-off curve between PICP and PI width illustrates the formulation balancing two objectives, with the lower left area being preferable, as it indicates high PICP and narrow PI width. The characteristics of this trade-off in the Gaussian dataset were chosen as an example because it contains both high- and low-volatility noise regions. As shown in \cref{fig:tradeoff_dgpgaussian}, any formulation can achieve any level of PICP by varying $\gamma$. The trade-off curve comparing PINAW and PICP shows that the characteristics of the CWC family are aligned and display comparable performance, even though our formulation being less favorable when assessed with PINAW. The benchmarked methods perform well in reducing PINAW because its loss function incorporates a PINAW term that corresponds with the evaluation criteria. Regarding the trade-off between PINALW and PICP, our formulation performs well at minimizing large PI width, as the loss function penalizes the $K$-largest PI widths more heavily. Among the CWC family, $\text{CWC}_{\text{Quan}}$ effectively reduces larger widths due to its two-norm characteristic that imposes a heavier penalty on larger PI widths. However, using linearly spaced $\gamma$, the trade-off curve in the CWC family reveals many closely positioned operating points. This is due to the mathematical formulation of the CWC family in which $\gamma$ is located inside the exponential term, affecting the penalization of deviations from PICP. Consequently, this makes it difficult to balance the two objectives and complicates the selection of an appropriate operating point. In formulations where the penalty parameters control two objectives in additive form, QD and Sum-$k$, a clear trade-off trend can be observed: increasing $\gamma$ directly decreases the PI width while losing the PICP. In contrast, increasing the trade-off parameter in the CWC family does not necessarily guarantee an enhancement in PICP. Additionally, the $\text{CWC}_{\text{Quan}}$ trade-off curve demonstrates a lack of smoothness, even after averaging 100 trials, with its trade-off trend remaining unclear.
	\begin{figure}[ht]
		\centering
			\includegraphics[width=0.58\linewidth]{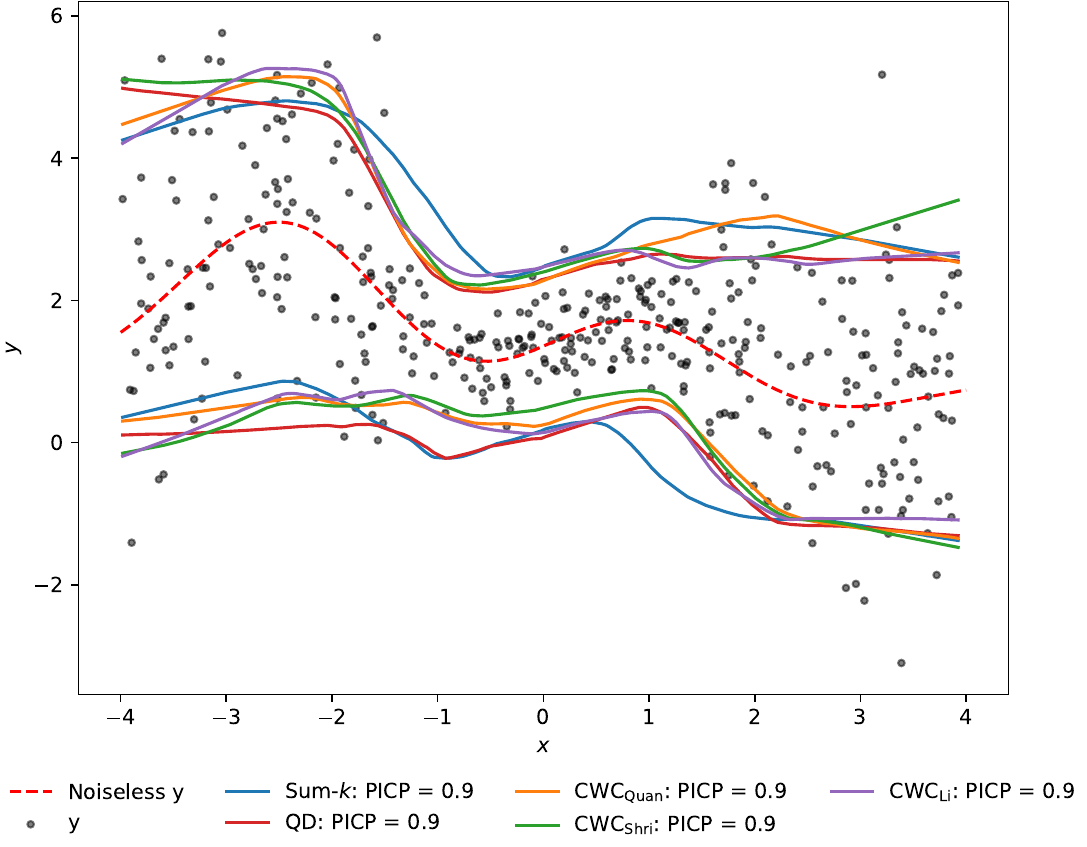}
			\caption{Comparison of PI characteristics from each formulation in the sum of Gaussian dataset.}
		\centering
		\label{fig:piresult_dgpgaussian}
	\end{figure}
\paragraph{Characteristics of PI.} The PI results for the various formulations are displayed in \cref{fig:piresult_dgpgaussian} with selecting equal PICP of 0.9. This figure illustrates that the sum of Gaussian data exhibits significant volatile noise when $\vert x \vert > 1.5$. Each formulation aims to capture the data point to attain the 0.9 PICP, where PI is large for the high variance region of the data and narrower for the low variance regions. Notably, in the highly volatile region, the sum-$k$ formulation maintains the smallest width compared to all other methods, although it displays a wider width in a low-volatility region. This outcome aligns with our objective of designing the formulation to minimize the large PI width in high-variance regions while permitting the PI in low-variance areas to be slightly wider than usual. In addition, our formulation changes the distribution of PI widths by aiming to decrease the large PI widths while allowing the smaller widths to increase. Thus, we will closely examine the PI width distribution across all methods to demonstrate how the formulation impacts PI widths.

\begin{figure}[ht]
		\centering
			\includegraphics[width=0.85\linewidth]{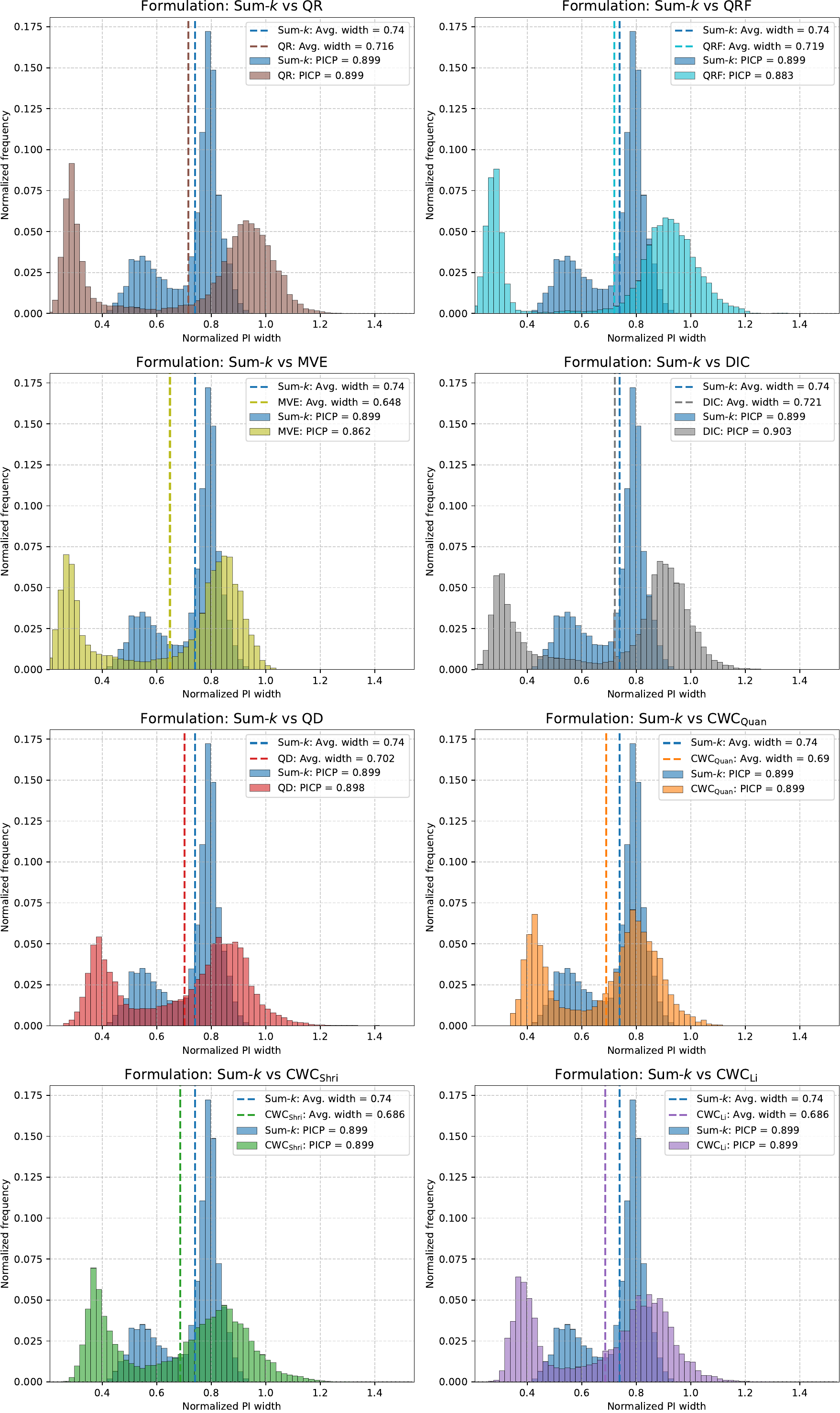}
			\caption{Comparison of the PI width histogram aggregated across 100 trials in the sum of the Gaussian dataset.}
		\centering
		\label{fig:hist_piwidth_dgpgaussian_exp2}
	\end{figure}

\paragraph{Distribution of PI width.}
The setting of an experiment to observe the distribution of PI width is described in \Cref{subsubsec:expdesign_controlpicp}. The histogram in \Cref{fig:hist_piwidth_dgpgaussian_exp2} illustrates the distribution of PI widths from our proposed formulation, compared with other methods, using the sum of Gaussian data as an example. The PI width histogram for all methods reveals two distinct peaks, indicating regions of high and low volatility that align with the DGP of the Gaussian dataset. Notably, the tail of the PI width distribution for the sum-$k$ formulation is the shortest compared to other methods, suggesting that, at the same PICP, our formulation effectively reduces the large PI width. However, while the larger PI widths decrease, the narrow PI width from our formulation tends to widen. Consequently, our formulation results in a PI width histogram with the least variation, showing a two-sided inward shift in distribution. This reduction in the variability of PI width is the effect of setting the hyperparameter $\lambda$ to  0.1, which penalizes the mean of large PI widths relative to the mean of narrow PI widths by a factor of ten. If $\lambda$ is increased, the PI width distribution widens and aligns with that observed when using PINAW as a loss function, which penalizes PI width equally. All synthetic data in this experiment exhibit these traits, indicating a two-sided inward shift in our PI width distribution. In summary, the hyperparameter $\lambda$ influences the PI width histogram as described, allowing the user to choose any $\lambda$ according to their specifications.

\paragraph{Performance metrics.} The performance metrics are reported as the averaging over all data trials with the standard deviation shown in \Cref{tab:experiment_performance_comparison}. The criterion for selecting the operating point for comparison is outlined in \Cref{subsubsec:expdesign_controlpicp}. \Cref{tab:experiment_performance_comparison} shows that our formulation generally achieves the lowest PINALW across most datasets when PICP is controlled. However, certain methods do not reach a PICP of 0.9 in some datasets. For instance, the $\text{CWC}_{\text{Quan}}$ cannot achieve PICP 0.9 in the case of the polynomial function, but results in the lowest PINALW, while our formulation remains close to 0.9 for PICP. In MVE, PICP regularly falls under 0.9, especially within multivariate functions, because the noise does not adhere to the Gaussian assumption required by MVE. Moreover, as the number of features rises, MVE's performance is adversely affected by the limited sample size in the loglikelihood formulation. The QR and QRF methods have no trade-off hyperparameters, so PICP aligns with the desired coverage probability by setting the 0.05 and 0.95 as lower and upper quantiles. Consequently, for multivariate functions, PICP from QR falls from 0.9 due to its failure to match the specific 0.05 and 0.95 quantiles.

When examining PI width metrics—specifically PINAW, PINALW, and the Winkler score—alongside a consistent PICP of approximately 0.9, the sum-$k$ formulation demonstrates the lowest PINALW across most datasets. However, the sum-$k$ formulation showed an increase in PINAW. To enhance PINAW, we can increase the parameter $\lambda$ in our formulation, which results in improved performance in PINAW but degrades PINALW, reflecting the specific requirements of each scenario. Regarding the Winkler score, QR-based methods excel in performance when evaluated, as this score is inherently related to the pinball loss function, which serves as their objective. A lower Winkler score indicates a closer alignment with quantiles 0.05 and 0.95. Between QR and QRF, QR shows a lower Winkler score due to the model's complexity. While QRF employs a tree-based model, QR utilizes a neural network with three hidden layers, resulting in superior performance from the more complex model. The lowest Winkler score indicates alignment with quantiles but does not guarantee the lowest PI width. This experiment confirms that PIs from PI-based loss with the lowest PINAW do not consistently align with a fixed quantile, as mentioned in \cite{Chen2024}.

In our analysis of training convergence with the multivariate function, we found that configuring the maximum epochs along with a patience parameter, as discussed in \Cref{subsec:tradeoffexp} for $\text{CWC}_{\text{Quan}}$, $\text{CWC}_{\text{Li}}$, and DIC loss, could experience difficulties in obtaining a convergence. When a low patience parameter and a low number of epochs are used, the PI width from these methods becomes excessively wide. The convergence process for these methods is notably slow, with thousands of iterations, while alternative methods can converge only hundreds of iterations. Additionally, a very high learning rate can cause loss divergence and instability. To address the divergence issue, we increase the maximum epochs and patience to facilitate convergence, with performance results displayed in \Cref{tab:experiment_performance_comparison}. A slow convergence of $\text{CWC}_{\text{Quan}}$ and $\text{CWC}_{\text{Li}}$ occurs from the mathematical formulation, which is multiplicative in the relationship between PI width and PICP term. Because it results in a PI width of zero at a global minimum, providing an undesired property for prediction intervals \cite{Pearce2018}. Therefore, an additive version of the loss function is more compatible with gradient-based algorithms, demonstrated by the usual convergence performance observed in $\text{CWC}_{\text{Shri}}$. Even with an additive term in DIC, the loss function experiences discontinuities, causing very high loss values when PICP is below 0.9 and dropping to decimal values once PICP exceeds 0.9. This requires a longer time to achieve low PI width while meeting PICP requirements due to the function's discontinuity. However, all CWC families were proposed using a heuristic optimization instead of a gradient-based one. The heuristic approach is employed because the original loss is non-differentiable, caused by the count function. In this experiment, we have already adopted the smooth approximation of the counting function for compatibility with gradient-based algorithms, but it still performs poorly in convergence. As a result, we decided to exclude $\text{CWC}_{\text{Quan}}$, $\text{CWC}_{\text{Li}}$, and DIC from the benchmark candidates in the real-world dataset.

\clearpage
\clearpage

\begin{table} 
\caption{Comparison of PIs performance indices on a validation set of synthesis datasets, maintaining a controlled PICP of 0.9 (mean $\pm$ one standard deviation). \textbf{The bold value} indicates the best performance, with an acceptable PICP that does not fall below 0.89.}

\begin{tabular}{p{2cm}p{1.8cm}p{2.8cm}p{2.8cm}p{2.8cm}p{3cm}} 
\hline 
Data & Method & PICP& PINAW& PINALW& Winkler\\ 
\hline
\multirow{8}{2cm}{Sum of Gaussian function} 
& QR  & 0.8990 $\pm$ \; 0.0134 & 0.7163 $\pm$ \; 0.0282 & 0.9751 $\pm$ \; 0.0427 & \bf{0.9051 $\pm$ 0.0495}\\ 
& QRF  & 0.8833 $\pm$ \; 0.0164 & 0.7192 $\pm$ \; 0.0164 & 0.9746 $\pm$ \; 0.0254 & 0.9547 $\pm$ \; 0.0442\\ 
& MVE  & 0.8616 $\pm$ \; 0.0229 & 0.6482 $\pm$ \; 0.0206 & 0.8671 $\pm$ \; 0.0312 & 0.9551 $\pm$ \; 0.0695\\ 
& DIC  & 0.9028 $\pm$ \; 0.0039 & 0.7211 $\pm$ \; 0.0296 & 0.9497 $\pm$ \; 0.0390 & 0.9361 $\pm$ \; 0.0520\\ 
& QD  & 0.8977 $\pm$ \; 0.0054 & 0.7022 $\pm$ \; 0.0249 & 0.8917 $\pm$ \; 0.0375 & 0.9841 $\pm$ \; 0.0600\\ 
& $\text{CWC}_{\text{Quan}}$  & 0.8994 $\pm$ \; 0.0067 & 0.6899 $\pm$ \; 0.0296 & 0.8475 $\pm$ \; 0.0399 & 0.9830 $\pm$ \; 0.0603\\ 
& $\text{CWC}_{\text{Shri}}$  & 0.8990 $\pm$ \; 0.0054 & \bf{0.6859 $\pm$ 0.0308} & 0.8845 $\pm$ \; 0.0489 & 0.9681 $\pm$ \; 0.0604\\ 
& $\text{CWC}_{\text{Li}}$  & 0.8992 $\pm$ \; 0.0056 & 0.6861 $\pm$ \; 0.0252 & 0.8803 $\pm$ \; 0.0380 & 0.9681 $\pm$ \; 0.0561\\ 
& Sum-$k$  & 0.8995 $\pm$ \; 0.0056 & 0.7396 $\pm$ \; 0.0269 & \bf{0.8168 $\pm$ 0.0328} & 1.0359 $\pm$ \; 0.0621\\
\hline
\multirow{8}{2cm}{Polynomial function} 
& QR  & 0.8967 $\pm$ \; 0.0233 & 0.4077 $\pm$ \; 0.0274 & 0.7230 $\pm$ \; 0.0517 & 0.5016 $\pm$ \; 0.0507\\ 
& QRF  & 0.8938 $\pm$ \; 0.0251 & 0.3516 $\pm$ \; 0.0194 & 0.6062 $\pm$ \; 0.0361 & \bf{0.4576 $\pm$  0.0507}\\ 
& MVE  & 0.8680 $\pm$ \; 0.0591 & 0.3573 $\pm$ \; 0.0309 & 0.6131 $\pm$ \; 0.0481 & 0.5869 $\pm$ \; 0.1685\\ 
& DIC  & 0.9037 $\pm$ \; 0.0036 & 0.3692 $\pm$ \; 0.0289 & 0.6379 $\pm$ \; 0.0529 & 0.5497 $\pm$ \; 0.0674\\ 
& QD  & 0.8968 $\pm$ \; 0.0061 & 0.3153 $\pm$ \; 0.0478 & 0.4827 $\pm$ \; 0.0665 & 0.7361 $\pm$ \; 0.1126\\ 
& $\text{CWC}_{\text{Quan}}$  & 0.8705 $\pm$ \; 0.0191 & 0.3366 $\pm$ \; 0.1229 & 0.4456 $\pm$ \; 0.1381 & 1.0408 $\pm$ \; 0.2040\\ 
& $\text{CWC}_{\text{Shri}}$  & 0.9005 $\pm$ \; 0.0048 & \bf{0.3028 $\pm$ 0.0280} & \bf{0.4625 $\pm$ 0.0516} & 0.7645 $\pm$ \; 0.1090\\ 
& $\text{CWC}_{\text{Li}}$  & 0.8863 $\pm$ \; 0.0141 & 0.4094 $\pm$ \; 0.1826 & 0.5650 $\pm$ \; 0.2087 & 0.9640 $\pm$ \; 0.2248\\ 
& Sum-$k$  & 0.8990 $\pm$ \; 0.0060 & 0.3707 $\pm$ \; 0.0344 & 0.4753 $\pm$ \; 0.0474 & 0.8813 $\pm$ \; 0.1049\\ 
\hline
\multirow{8}{2cm}{Sinusoid function} 
& QR  & 0.9041 $\pm$ \; 0.0201 & 0.5737 $\pm$ \; 0.0242 & 0.7806 $\pm$ \; 0.0352 & \bf{0.7081 $\pm$ 0.0443}\\ 
& QRF  & 0.8832 $\pm$ \; 0.0221 & 0.6071 $\pm$ \; 0.0198 & 0.8171 $\pm$ \; 0.0342 & 0.8220 $\pm$ \; 0.0556\\ 
& MVE  & 0.8722 $\pm$ \; 0.0290 & 0.5263 $\pm$ \; 0.0195 & 0.7074 $\pm$ \; 0.0290 & 0.7356 $\pm$ \; 0.0571\\ 
& DIC  & 0.9038 $\pm$ \; 0.0054 & 0.6167 $\pm$ \; 0.0312 & 0.8080 $\pm$ \; 0.0437 & 0.7878 $\pm$ \; 0.0518\\ 
& QD  & 0.8954 $\pm$ \; 0.0091 & 0.5827 $\pm$ \; 0.0252 & 0.7140 $\pm$ \; 0.0335 & 0.8208 $\pm$ \; 0.0689\\ 
& $\text{CWC}_{\text{Quan}}$  & 0.9002 $\pm$ \; 0.0079 & 0.5532 $\pm$ \; 0.0265 & 0.6526 $\pm$ \; 0.0332 & 0.7984 $\pm$ \; 0.0653\\ 
& $\text{CWC}_{\text{Shri}}$  & 0.8988 $\pm$ \; 0.0077 & \bf{0.5463 $\pm$ 0.0215} & 0.6715 $\pm$ \; 0.0285 & 0.7803 $\pm$ \; 0.0589\\ 
& $\text{CWC}_{\text{Li}}$  & 0.9001 $\pm$ \; 0.0091 & 0.5523 $\pm$ \; 0.0237 & 0.6802 $\pm$ \; 0.0333 & 0.7873 $\pm$ \; 0.0593\\ 
& Sum-$k$  & 0.8995 $\pm$ \; 0.0096 & 0.6012 $\pm$ \; 0.0235 & \bf{0.6226 $\pm$ 0.0240} & 0.8737 $\pm$ \; 0.0674\\ 
\hline
\multirow{9}{2cm}{Multivariate function} 
& QR  & 0.8160 $\pm$ \; 0.0295 & 0.5517 $\pm$ \; 0.0268 & 0.6722 $\pm$ \; 0.0350 & 0.8990 $\pm$ \; 0.0746\\ 
& QRF  & 0.9123 $\pm$ \; 0.0220 & 0.7534 $\pm$ \; 0.0184 & 0.8471 $\pm$ \; 0.0216 & 0.9350 $\pm$ \; 0.0616\\ 
& MVE  & 0.4705 $\pm$ \; 0.0401 & 0.2981 $\pm$ \; 0.0183 & 0.3669 $\pm$ \; 0.0264 & 1.8691 $\pm$ \; 0.1623\\ 
& DIC  & 0.9050 $\pm$ \; 0.0056 & 0.6800 $\pm$ \; 0.0435 & 0.8249 $\pm$ \; 0.0524 & 0.8784 $\pm$ \; 0.0629\\ 
& QD  & 0.8880 $\pm$ \; 0.0073 & 0.7066 $\pm$ \; 0.0878 & 0.8414 $\pm$ \; 0.1156 & 0.9829 $\pm$ \; 0.1065\\ 
& $\text{CWC}_{\text{Quan}}$  & 0.8915 $\pm$ \; 0.0111 & 0.8353 $\pm$ \; 0.1185 & 1.0006 $\pm$ \; 0.1784 & 1.1242 $\pm$ \; 0.1415\\ 
& $\text{CWC}_{\text{Shri}}$  & 0.8922 $\pm$ \; 0.0121 & 0.7751 $\pm$ \; 0.0711 & 0.9484 $\pm$ \; 0.0982 & 1.0500 $\pm$ \; 0.0988\\ 
& $\text{CWC}_{\text{Li}}$  & 0.8674 $\pm$ \; 0.0255 & 0.7269 $\pm$ \; 0.0555 & 0.8890 $\pm$ \; 0.0744 & 1.1013 $\pm$ \; 0.1041\\ 
& Sum-$k$  & 0.9008 $\pm$ \; 0.0055 & \bf{0.6407 $\pm$ 0.0395} & \bf{0.6792 $\pm$ 0.0486} & \bf{0.8798 $\pm$ 0.0656}\\ 
\hline
\end{tabular} 
\label{tab:experiment_performance_comparison}
\end{table}

\clearpage
\section{Experiment on solar forecasting} \label{sec:exponrealdataset}
\paragraph{Forecasting specification} We aim to deliver the one-hour-ahead solar irradiance forecast PI  from 07:00 to 17:00 with a confidence level of 0.9 at a 15-minute resolution, corresponding to 4 lead times. This forecasted PI indicates the uncertainty in solar energy, assisting system operators in decision-making for better reserve preparation and generation planning.

\subsection*{Dataset} 
All predictors used in this experiment can be divided into two groups: lagged regressors and future regressors. The lagged regressors use the lag of measurement values as a predictor, classified into auto-lagged and exogenous lagged regressors. In this study, solar irradiance is the auto-lagged regressor, while cloud data is the exogenous lagged regressor. The future regressor refers to the exogenous variable assumed to be available for future periods. In this study, we utilized clear-sky irradiance, Numerical Weather Prediction (NWP) forecast data, and hour index as future regressors. The full description of all predictors is provided below.

\paragraph{Lagged regressors.} For lagged regressors, solar irradiance historical data ($I$) is collected from solar stations provided by the Department of Alternative Energy Development and Efficiency (DeDe), Ministry of Energy, Thailand, which includes ten solar sites across various provinces in Central Thailand, spanning from January to December 2023, with the 15-minute resolution. The cloud data is extracted from RGB cloud images sourced from the Himawari-8 satellite covering Thailand, which has a spatial resolution of $2 \times 2 \; \text{km}^{2}$. The cloud images are collected between 06:00 and 19:50 at 10-minute intervals and resampled to align with specifications. The cloud index was calculated using the formula $\ci = \frac{X - \text{LB}}{\text{UB} - \text{LB}}$, where $X$ represents the pixel intensity, LB and UB denote the lower and upper bounds set at 0 and 255, respectively. We utilized the R-channel cloud index denoted as $\ci_{R}$ due to its strong correlation from pre-analysis. As a result, we used a four-period lag for solar irradiance and cloud index: $t-45, t-30, t-15, t$, depending on the available cloud data and the forecast times specified, where $t$ represents the timestamp in minutes.

\paragraph{Future regressors.} We generated clear-sky irradiance, denoted as $\iclr$ from Ineichen clear-sky model \cite{Ineichen2002} where the Linke turbidity component in the clear-sky model was calculated monthly using PVlib \cite{Anderson2023}. Next, we imported re-analyzed forecast data from MERRA-2 (Modern-Era Retrospective Analysis for Research and Applications) and downloaded it from the Solar Radiation Data Service (SoDa) \url{https://www.soda-pro.com}. The forecast data originates from the Global Forecast System (GFS) weather model developed by the National Centers for Environmental Prediction (NCEP), based on NWP with the 15-minute resolution. The data includes temperature, relative humidity, pressure, wind speed and direction, rainfall, snowfall, snow depth, and short-wave irradiance. In this study, we employed the short-wave radiation from NCEP, denoted as $\inwp$, to provide the solar irradiance forecast due to a strong correlation. Additionally, we also include the hour index (HI) as a future regressor to represent the hour of the day. The target variables are set as $I(t+15), I(t+30), I(t+45), I(t+60)$ according to the lead time specifications, where future regressors are aligned with the forecast times $t+15, t+30, t+45, t+60$.

As a result, the dataset samples were spanned from 06:45 to 17:00. For each sample, the clear-sky index was calculated as $k=I/\iclr$ and averaged daily as $\overline{k}$. Sky conditions were classified into clear, partly cloudy, and cloudy. Clear-sky data was detected on the smoothness of irradiance, while the remaining conditions were classified using $\overline{k}$: cloudy if $\overline{k} < 0.75$, otherwise partly cloudy. After classifying sky conditions, the dataset comprises clear-sky, partly cloudy, and cloudy conditions in a ratio of $52:11:37$. It is split into training, validation, and test sets using an $80:10:10$ ratio, ensuring equal sky condition representation. This division yields 91,055 training samples, 11,455 validation samples, and 11,283 test samples.

\subsection*{Model and loss function} \label{sec:solar_model}
\begin{figure}[ht]
	\centering
		\includegraphics[width=1\linewidth]{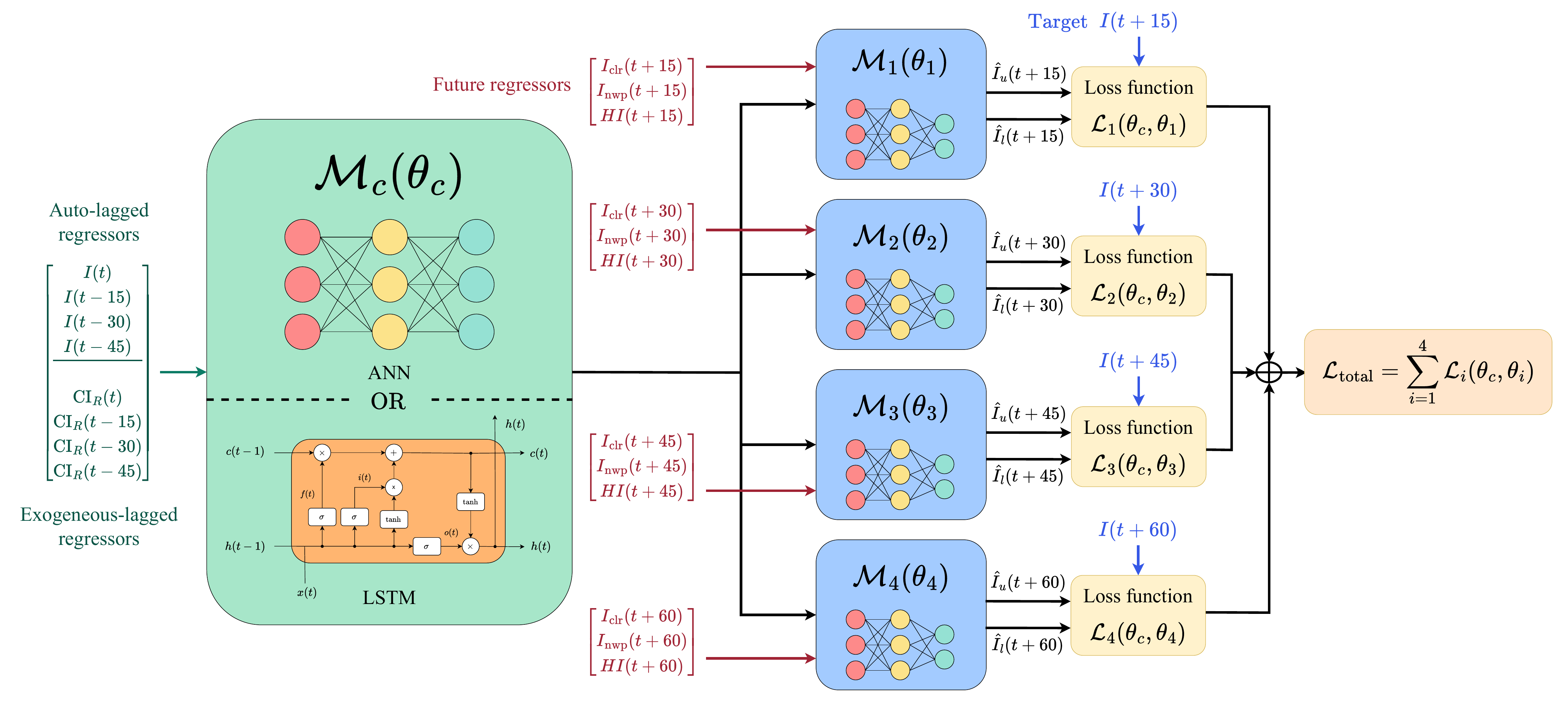}
		\caption{The NN model architecture used in the solar data experiment includes a common model $\mathcal{M}_{c}$ with received lagged regressors as inputs and a submodel $\mathcal{M}_{i}$ with received future regressors, where the PI outputs with the target are used to evaluate the loss function.}
	\centering
	\label{fig:solarmodel}
\end{figure}

We designed the model to release the PI for all forecasting lead times at once, as illustrated in 
\cref{fig:solarmodel}. The model has eight output neurons representing the upper and lower bounds for four-step forecasting. The model is designed to establish distinct predictors for each forecasting lead time, aligning them with the forecasted time. Therefore, the model consists of a common model and four separate lead-time-specific submodels. The common model, denoted as $\mathcal{M}_{c}$, handles lagged regressors, including solar irradiance measurements and cloud index. This common model serves as a shared input layer for all lead times. In this experiment, we explored two options for $\mathcal{M}_{c}$: an ANN and an LSTM architecture to capture the dynamics of the lagged features effectively. For ANN, we configured the hidden layers of $\mathcal{M}_{c}$ to have two layers, each with 100 neurons. Similarly, for the LSTM, we implemented two layers of LSTM cells, setting the hidden size of each cell to 45, which aligns the number of trainable parameters with that of the ANN for performance comparison. The second component of the model is referred to as separated lead time submodels, denoted as $\mathcal{M}_{i}$ for $i = 1, 2, \cdots, H$ where $H$ denotes the number of forecasting steps. Each $\mathcal{M}_{i}$ produces the upper and lower bound for the $i^{\mathrm{th}}$ step ahead prediction. Future regressors, including $\iclr, \inwp$, and HI, are integrated at the corresponding timestamps to the target variables shown in \cref{fig:solarmodel} because they directly explain the target variable at those timestamps. Each submodel comprises two hidden layers of an ANN, each containing 100 neurons and utilizing the ReLU activation function. Additionally, batch normalization layers are applied before the activation function to enhance training stability. These design choices result in 95,808 trainable parameters for the ANN and 99,278 for the LSTM. Consequently, the PI for each lead time, along with the corresponding target variables, is used to compute the loss function for each lead time ($\mathcal{L}_{i}$) as illustrated in \cref{fig:solarmodel}. The overall loss ($\mathcal{L}_{\text{total}}$) for training the model is then obtained by summing these individual losses into a single total loss, expressed as:
\begin{equation}
	\mathcal{L}_{\text{total}}(\theta_{c}, \theta_{1}, \theta_{2}, \theta_{3}, \theta_{4}) = \sum_{i=1}^{4} \mathcal{L}_{i}(\theta_{c}, \theta_{i}),
\end{equation}
where $\theta_{c}$ is the parameters of the common model, and $\theta_{i}$ for $i = 1, 2, 3, 4$ corresponds to the parameters of the submodel for the $i^{\mathrm{th}}$ lead time.

\subsection*{Experiment setting}
The hyperparameters $\gamma$ of QD, $\text{CWC}_\text{Shri}$, and the sum-$k$ formulation are tuned to reach a 0.9 PICP across all forecast horizons in the validation set, after which performance is evaluated on the test set. For the sum-$k$ formulation, the hyperparameter $k$ was set constant at 0.3, while $\lambda$ varies; we select $\lambda = 0.9$ because the PICP is acceptable and the large width is sufficiently small. For QR, we set the lower and upper quantiles at 0.05 and 0.95, respectively, to achieve an overall PICP of 0.9. In comparison to other benchmarked losses, we selected the ANN as the common model with the same architecture. For the sum-$k$ formulation, we applied both ANN and LSTM options in the common model to assess the model's effect while the submodel architecture remains the same across all methods. The Adam optimizer is used for all methods, with a batch size of $0.3N_{\text{training}}$. The maximum number of epochs is 2,000, with a patience of 100 for early stopping. The learning rate is fine-tuned by monitoring the loss during training to determine the optimal value for convergence.

\subsection*{Result and discussion} 
\Cref{tab:experimentsolar_performance_comparison} presents a comparison of the PICP, PINAW, PINALW, and Winkler scores of the test set for the QR, QD, $\text{CWC}_{\text{Shri}}$, and sum-$k$ models using both ANN and LSTM. The result on PICP shows that the methods required for tuning the hyperparameter can achieve PICP at a desired value across all lead times, whereas the PICP for QR decreases at a 30-minute lead time. The formulation with a trade-off parameter offers greater flexibility in choosing the operating point that yields the desired PICP. In contrast, setting fixed lower and upper quantiles for QR does not guarantee achieving the desired PICP when aggregating losses from different lead times. In the PI width aspect, when compared among the same ANN architecture with different loss functions, \cref{tab:experimentsolar_performance_comparison} shows that the sum-$k$ formulation significantly reduced the large PI width indicated by the lowest PINALW for all lead times, where the PINAW was also the narrowest at 15 and 45 minutes ahead. This decrease in PINALW corresponds with our design in the PI width function for the sum-$k$ formulation, imposing a greater penalty on larger PI widths. The Winkler score shows that the sum-$k$ formulation yields the highest score. This result indicates that the lower and upper bounds of the sum-$k$ formulation deviate the most from the 0.05 and 0.95 quantiles. The outcome aligns with the findings from the synthetic data, suggesting that the lowest PI width can be achieved without aligning the PI to a specific quantile. In practical scenarios like solar and wind forecasting, it is not necessary for the PI to match a fixed quantile; instead, the focus is primarily on the PI width.

\begin{figure}[ht]
	\centering
		\includegraphics[width=1\linewidth]{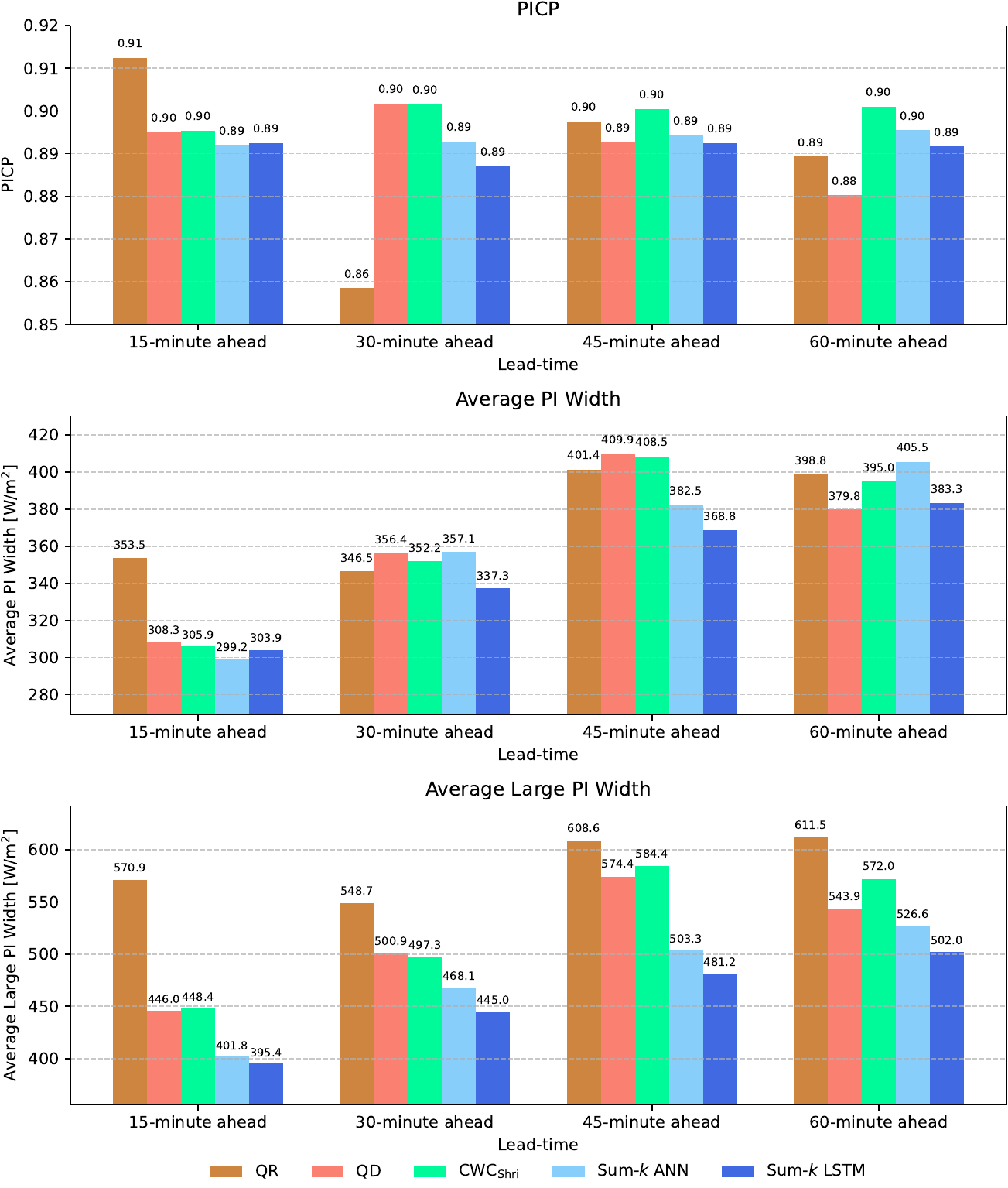}
		\caption{PI evaluation on the test set in the unit of $\wm$.}
	\centering
	\label{fig:bargraph_performance_multihorizon}
\end{figure}

\Cref{fig:bargraph_performance_multihorizon} illustrates a performance comparison on the test set regarding solar irradiance across all lead time in the unit of $\wm$. The bar graph showed that all methods can almost achieve PICP at the desired value across all lead times, except for QR. Our formulation achieves the lowest PINALW, offering a significant advantage over other methods while demonstrating good performance in terms of PINAW. As the number of forecast steps increases, the PI widths are generally wider due to the inherent variability of solar irradiance over a longer lead time. The results from this experiment show that the PI width from the sum-$k$ formulation aligns with this trend, demonstrating an increased PI width as the forecast lead time increases. However, this trend is not guaranteed for all methods, especially when all lead times are trained as an aggregated single loss function, as shown in the PI width between 45-minute and 60-minute lead times in \cref{fig:bargraph_performance_multihorizon}. To minimize the overall loss, the model may prioritize reducing the PI width in the longest lead time, which is often the widest PI width due to its inherent nature. 

\begin{figure}[ht]
	\centering
		\includegraphics[width=1\linewidth]{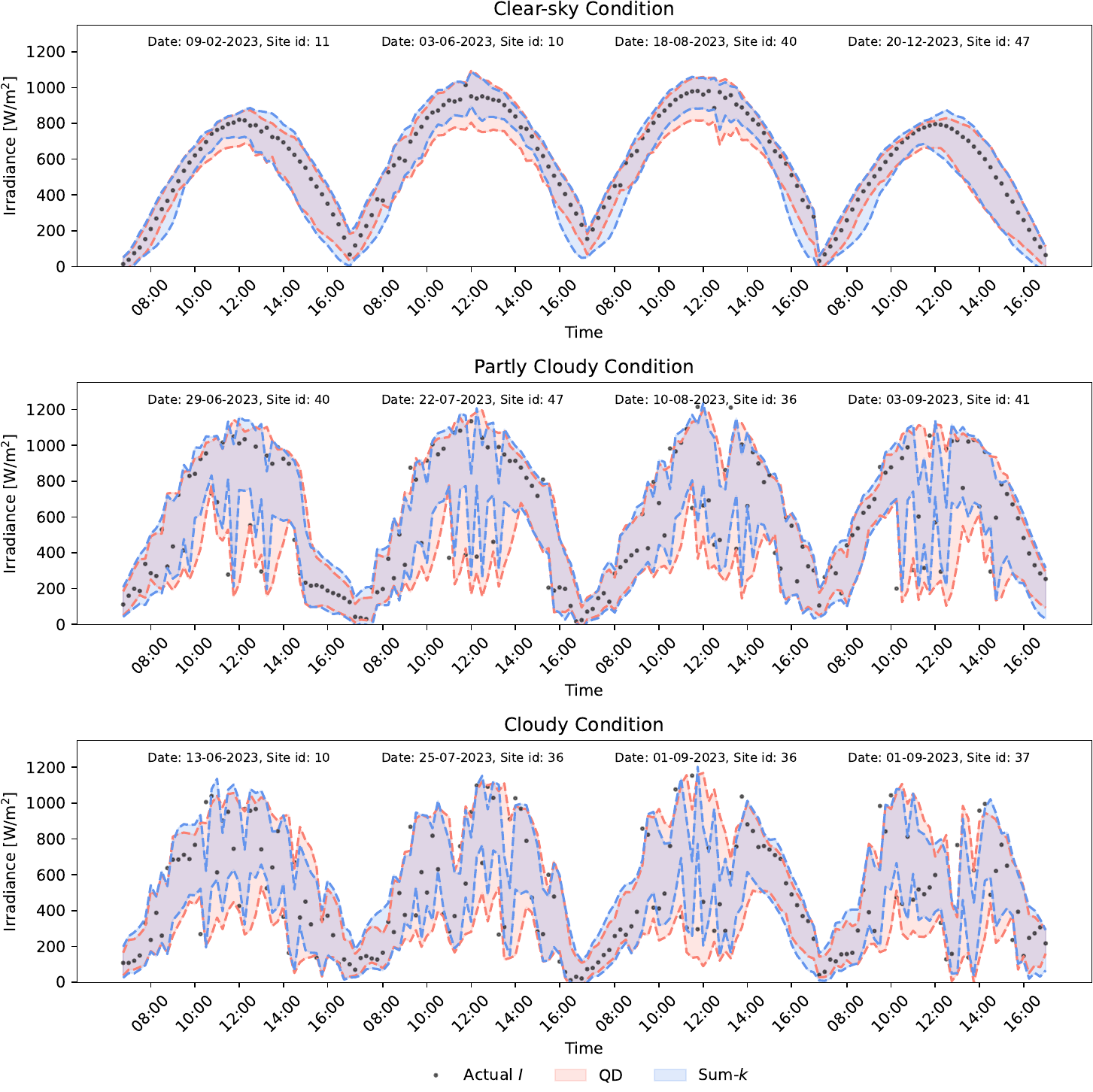}
		\caption{A comparison of the 15-minute ahead PI forecast of solar irradiance between \textbf{Sum-$k$ ANN and QD}.}
	\centering
	\label{fig:piplot_solar_15minahead_qdsumk}
\end{figure}

\Cref{fig:piplot_solar_15minahead_qdsumk} shows a time series plot of solar irradiance with 15-minute ahead PI s, comparing the QD and sum-$k$ formulations, separately plotted based on the sky condition. For each sky condition, we selected the first four dates where QD produced the largest PINAW to analyze and compare PI characteristics. It can be noticed that the sum-$k$ formulation significantly decreases the PI width during partly cloudy and cloudy conditions, which exhibit high volatility noise while keeping the PICP at approximately 0.9. In addition, with a suitable choice of $\lambda$ in the sum-$k$ formulation, the PI width from our formulation under clear-sky conditions performs well even in the low volatile noise case compared to the QD. Especially in this condition, it can be observed that the PI width from our formulation is narrower than the QD at noon while being wider than the QD in the morning and evening. This is because the sum-$k$ formulation provides a narrower PI width during periods of high uncertainty, typically at noon, and a slightly wider PI width during the morning and evening when uncertainty is lower.

\begin{figure}[ht]
	\centering
		\includegraphics[width=1\linewidth]{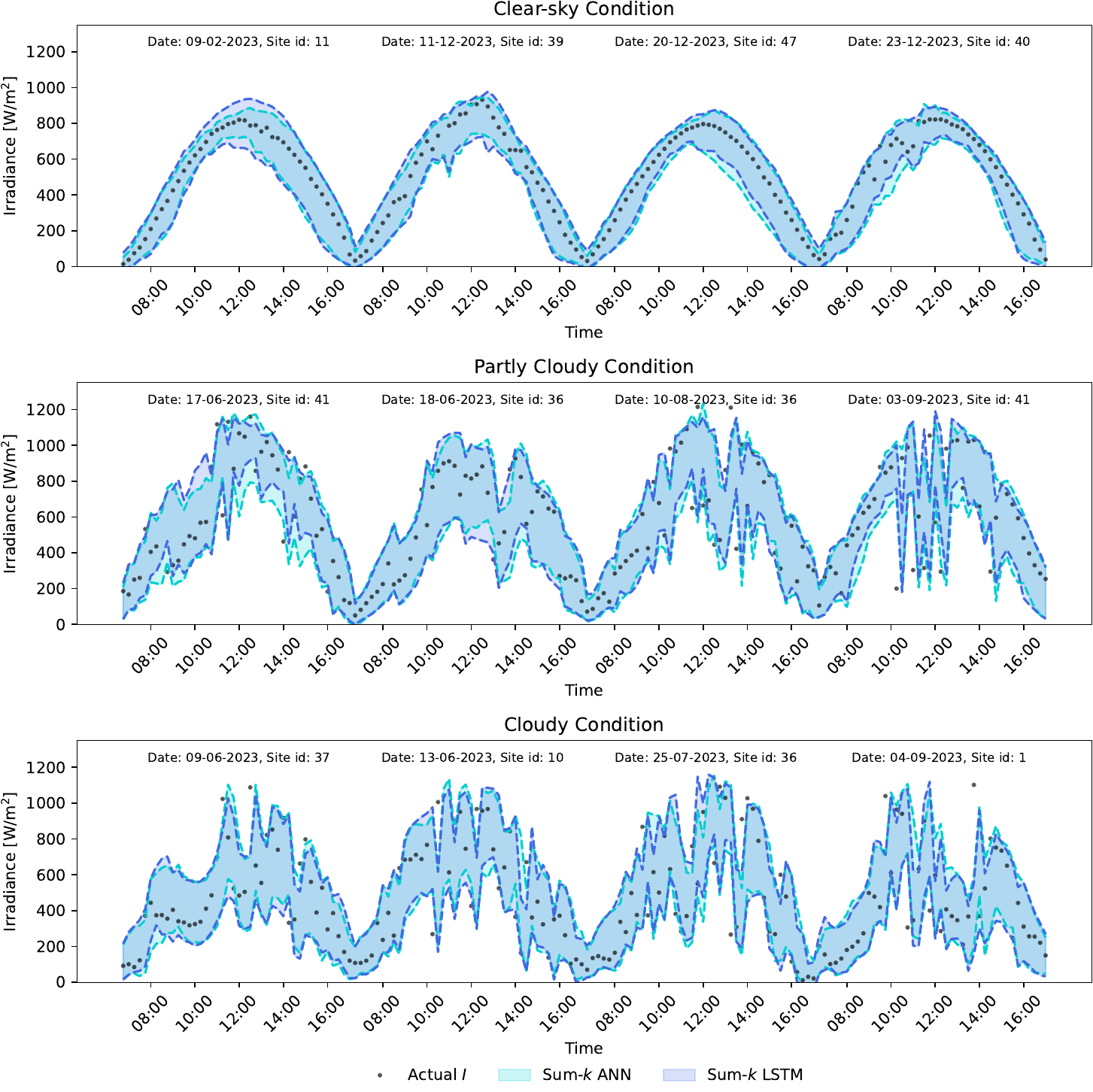}
		\caption{A comparison of the 15-minute ahead PI forecast of solar irradiance between \textbf{Sum-$k$ ANN and Sum-$k$ LSTM}.}
	\centering
	\label{fig:piplot_solar_15minahead_annlstm}
\end{figure}

\Cref{tab:experimentsolar_performance_comparison} also presented the performance comparison between the ANN and LSTM models. The LSTM model achieves the lowest PINALW at all lead times, also showing improvement in PINAW compared to the ANN model. As shown in \cref{fig:bargraph_performance_multihorizon}, the LSTM shows improved PINALW at all lead times when compared to the ANN, supported by the lower validation loss of the sum-$k$ formulation, which reflects a better ability to reduce the large PI width. In \cref{fig:piplot_solar_15minahead_annlstm}, the 15-minute-ahead PIs of solar irradiance time series are compared between the ANN and LSTM models, which utilize the same sum-$k$ loss function. We selected the first four dates on which the PI from the ANN model had the highest PINAW for each sky condition to observe the characteristics of the PI from LSTM on the dates when the ANN performed poorly. The results indicate that the PI characteristics of the two models are relatively similar. While the LSTM sometimes produces narrower or wider PI widths than the ANN on specific dates, it is notable that when the ANN generates a large PI width, the LSTM can slightly reduce it. The overall shapes of the PIs from both models are comparable due to the same loss function, though their performance differs slightly based on the inherent characteristics of the model. In conclusion, this experiment demonstrates that our formulation is suitable for various complex NN models trained with gradient-based algorithms, allowing the integration of more advanced architectures to enhance PI performance.

\begin{figure}[ht]
	\centering
	\begin{subfigure}[t]{1\textwidth}
		\centering
		\includegraphics[width=1\linewidth]{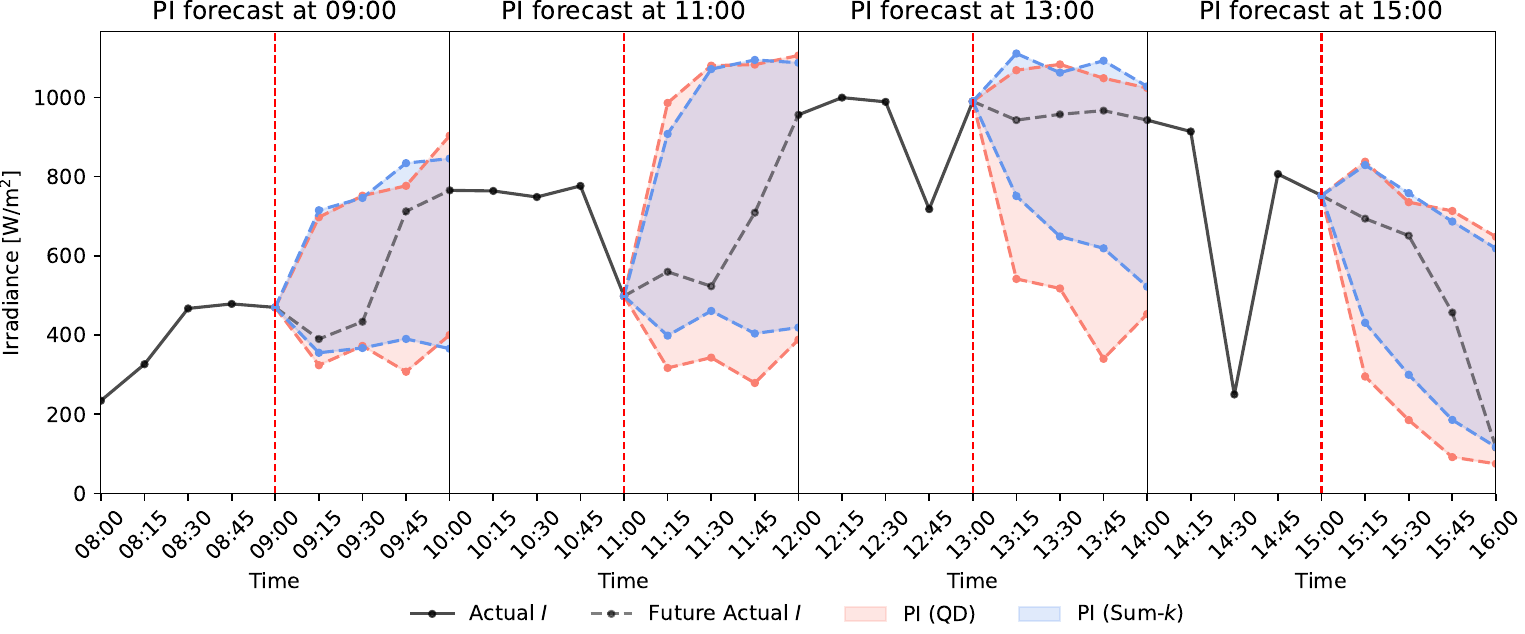}
		\caption{Actual future are covered PI.}
		\label{subfig:piplot_goodcond}
	\vfill
	\end{subfigure}
	\begin{subfigure}[t]{1\textwidth}
		\centering
		\includegraphics[width=1\linewidth]{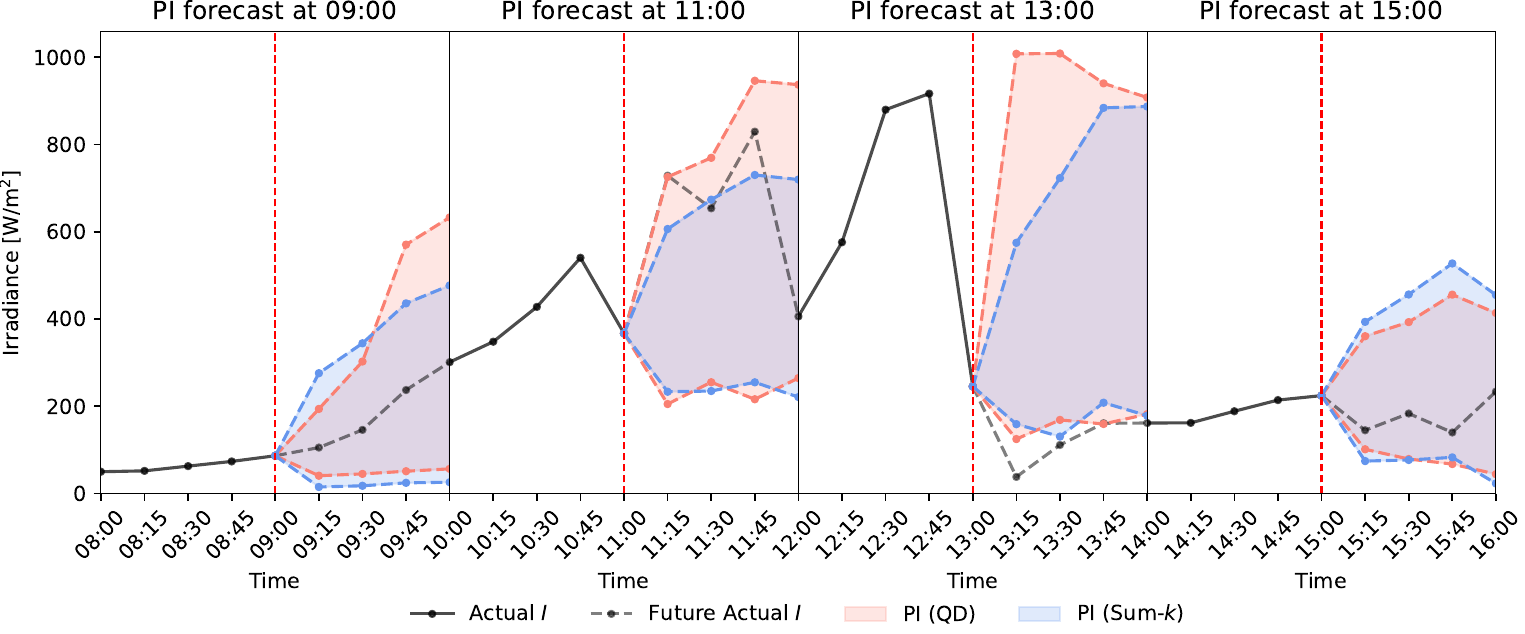}
		\caption{Actual future are not covered PI.}
		\label{subfig:piplot_badcond}
	\vfill
	\end{subfigure}
	\caption{A comparison between the Sum-$k$ and QD formulations of a real-time 4-step PI forecast for solar irradiance, released at varying times .}
	\label{fig:piplot_solar_60minutesahead_horizontal}
\end{figure}

To illustrate a real-time forecast, the 4-step ahead PIs are released during a specified time, as illustrated in \cref{fig:piplot_solar_60minutesahead_horizontal}. The PIs are compared between QD and the sum-$k$ formulation using the ANN model. \cref{subfig:piplot_goodcond} shows instances where the PIs cover the future actual irradiance for all lead times. In addition, our formulation shows a narrower PI width compared to the QD when the selected date involves a degree of uncertainty. As the PI characteristics establish upper and lower limits at a 0.9 confidence level, there is a 0.1 probability that the actual future irradiance could fall outside the PI, as observed on some dates illustrated in \cref{subfig:piplot_badcond}. Such behavior is typical, as the PI estimation can only ensure reliability at the user-specified confidence level. In conclusion, the PI from the sum-$k$ formulation successfully achieved the reliability specification as the PICP met the required confidence level. In addition, the reduction in the large PI width of solar irradiance is our advantage over the other formulation, which leads to a decrease in the operational cost for reserve power preparation.

\begin{table} 
\centering
\caption{Comparison of PI evaluation metrics on the test set of one-hour-ahead solar irradiance forecasting, with a resolution of 15 minutes and controlled PICP at 0.9 in the validation set. \textbf{The bold value} indicates the best performance, with an acceptable PICP}
\begin{tabular}{|p{2.2cm}p{1.2cm}p{1.2cm}p{1.2cm}p{1.2cm}|} 
\hline
\rowcolor{CornflowerBlue!50}
\multicolumn{5}{|c|}{15-minute ahead} \\ 
\hline
Method  & PICP & PINAW & PINALW & Winkler\\ 
\hline
QR & 0.912 & 0.395 & 0.638 & \bf{0.484} \\ 
QD & 0.895 & 0.345 & 0.499 & 0.572 \\ 
$\text{CWC}_{\text{Shri}}$ & 0.895 & 0.342 & 0.501 & 0.611 \\ 
Sum-$k$ ANN & 0.892 & \bf{0.335} & 0.449 & 0.656 \\ 
Sum-$k$ LSTM & 0.892 & 0.340 & \bf{0.442} & 0.675 \\ 
\hline
\rowcolor{SeaGreen!50}
\multicolumn{5}{|c|}{30-minute ahead} \\ 
\hline
Method  & PICP & PINAW & PINALW & Winkler\\ 
\hline
QR & 0.859 & 0.388 & 0.614 & \bf{0.547} \\ 
QD & 0.902 & 0.399 & 0.560 & 0.627 \\ 
$\text{CWC}_{\text{Shri}}$ & 0.902 & 0.394 & 0.556 & 0.647 \\ 
Sum-$k$ ANN & 0.893 & 0.399 & 0.523 & 0.694 \\ 
Sum-$k$ LSTM & 0.887 & \bf{0.377} & \bf{0.498} & 0.666 \\ 
\hline
\rowcolor{Apricot!50}
\multicolumn{5}{|c|}{45-minute ahead} \\ 
\hline
Method  & PICP & PINAW & PINALW & Winkler\\ 
\hline
QR & 0.898 & 0.449 & 0.681 & \bf{0.569} \\ 
QD & 0.893 & 0.458 & 0.642 & 0.644 \\ 
$\text{CWC}_{\text{Shri}}$ & 0.900 & 0.457 & 0.653 & 0.643 \\ 
Sum-$k$ ANN & 0.894 & 0.428 & 0.563 & 0.716 \\ 
Sum-$k$ LSTM & 0.892 & \bf{0.412} & \bf{0.538} & 0.694 \\ 
\hline
\rowcolor{VioletRed!50}
\multicolumn{5}{|c|}{60-minute ahead} \\ 
\hline
Method  & PICP & PINAW & PINALW & Winkler\\ 
\hline
QR & 0.889 & 0.446 & 0.684 & \bf{0.579} \\ 
QD & 0.880 & \bf{0.425} & 0.608 & 0.676 \\ 
$\text{CWC}_{\text{Shri}}$ & 0.901 & 0.442 & 0.640 & 0.684 \\ 
Sum-$k$ ANN & 0.896 & 0.454 & 0.589 & 0.704 \\ 
Sum-$k$ LSTM & 0.892 & 0.429 & \bf{0.561} & 0.713 \\ 
\hline
\end{tabular} 
\label{tab:experimentsolar_performance_comparison}
\end{table}

\clearpage
\section{Conclusion} 
This article proposes a novel PI-based loss function, the sum-$k$ loss, which effectively reduces the large PI widths while maintaining the target coverage probability in the PI estimation framework. The PI width term in the sum-$k$ formulation heavily penalizes the average of $K$-largest width, introducing the reduction on the large PI width. As the PICP in the sum-$k$ loss involves calculating the non-differentiable count function, a new $\tanh$-smooth approximation is used to approximate the count function using a differentiable function. Thus, the proposed formulation is compatible with a gradient-based algorithm, allowing the use of complex NN architectures to capture the nonlinear characteristics of the data. The synthetic data results demonstrated that our loss function effectively reduced the large PI widths, shifting them to be lower than the PIs from other methods, evidently shown from the histogram of PI width. The performance indices from 100 trials of synthetic data confirmed that our formulation achieved the lowest PINALW in most datasets compared to benchmarked methods while maintaining the desired PICP. However, our average PI width was slightly higher than the other methods with the same PICP. In the solar forecasting experiment, our loss function was applied to provide a 1-hour ahead with a 4-step prediction interval forecasting. The results showed that our methods yielded an average PI width comparable to other approaches but significantly reduced the PI width when the data exhibited high volatility. Furthermore, the experiment demonstrated that the proposed method was suitable for the LSTM model, known for its effectiveness in capturing temporal dynamics in time-series data. The benefit of decreasing the large PI width in solar applications from our work lowers the operational cost of preparing the reserve power for economic dispatch and the unit commitment problem. The PI from our formulation avoids conservative solutions for unit commitment or economic dispatch when decision-making is based on the worst-case scenario reflected in the large PI width. The proposed loss function can be applied beyond renewable forecasting to provide PIs in various applications, helping to reduce large PI widths that may result from undesirable over-estimations of uncertainty. Furthermore, various deep-learning techniques can also be integrated into our PI construction framework. However, a limitation of our approach is that the PI width for low-volatile data tends to be slightly broader than that of other methods, which results in an overall increase in average PI width. Additionally, fine-tuning the appropriate value of $\gamma$ to achieve the desired PICP requires multiple neural network training, leading to high computational costs. To tackle this challenge, multi-task learning, a multi-objective optimization approach, can be employed to automatically adjust the trade-off parameter during training, thereby optimizing the two conflicting objectives more efficiently.

\section{Acknowledgment}
The first author expresses gratitude for the CUEE High-Performance Scholarship from the Department of Electrical Engineering at Chulalongkorn University, which provided research funding and study support throughout his Master program. This work is partially supported by Chula Engineering Research Grant 2024. We also acknowledge the support from the Division of Solar Energy Development, Department of Alternative Energy Development and Efficiency (DeDe), Ministry of Energy, Thailand, for providing the historical solar irradiance measurement data utilized in this research.

\section{References}
\bibliography{ref.bib}
\bibliographystyle{alphaabbr}

\end{document}